\documentclass[runningheads]{llncs}

\usepackage{accv}

\usepackage{accvabbrv}
\usepackage{graphicx}
\usepackage{adjustbox}
\usepackage{booktabs}
\usepackage{subcaption}
\usepackage{pgfplots}
\pgfplotsset{compat=1.17}
\usepgfplotslibrary{fillbetween}
\usepgfplotslibrary{groupplots}
\providecolor{ColorGAUsel}{HTML}{42D4F4}
\providecolor{ColorGEMsel}{HTML}{808000}
\providecolor{ColorGaU}{HTML}{42D4F4}
\providecolor{ColorGemanMcClure}{HTML}{808000}
\providecolor{ColorIGLO}{HTML}{1A7A2E}
\providecolor{ColorIGLZ}{HTML}{FD8D3C}
\providecolor{ColorIGLift}{HTML}{41AB5D}
\providecolor{ColorIGLiftECM}{HTML}{66C2A5}
\providecolor{ColorIGMLE}{HTML}{4363D8}
\providecolor{ColorIGMLEIRLSStu}{HTML}{4363D8}
\providecolor{ColorIGPoseLib}{HTML}{FDAE6B}
\providecolor{ColorIGReGeMM}{HTML}{8C564B}
\providecolor{ColorIGRobust}{HTML}{3CB44B}
\providecolor{ColorIGRobustIRLSStu}{HTML}{3CB44B}
\providecolor{ColorIGTLSadapt}{HTML}{1F77B4}
\providecolor{ColorIGWIrls}{HTML}{A1D99B}
\providecolor{ColorIGsel}{HTML}{3CB44B}
\providecolor{ColorIGtFull}{HTML}{6A3D9A}
\providecolor{ColorJeffreys}{HTML}{E6194B}
\providecolor{ColorJeffreysIRLSStu}{HTML}{E6194B}
\providecolor{ColorMAGSAC}{HTML}{F58231}
\providecolor{ColorMAGSACsel}{HTML}{9A6324}
\providecolor{ColorMSAC}{HTML}{FFE119}
\providecolor{ColorMSACLMtrunc}{HTML}{FFE119}
\providecolor{ColorMSACLO}{HTML}{B5530B}
\providecolor{ColorMSACLZ}{HTML}{D62728}
\providecolor{ColorMSACPoseLib}{HTML}{B5530B}
\providecolor{ColorMSACsel}{HTML}{F58231}
\providecolor{ColorOracle}{HTML}{808080}
\providecolor{ColorRANSAC}{HTML}{FFB6C1}
\providecolor{ColorRANSACsel}{HTML}{FFB6C1}
\providecolor{ColorSIMFITsel}{HTML}{000075}
\providecolor{ColorWELsel}{HTML}{911EB4}
\providecolor{ColorWMGsel}{HTML}{F032E6}
\providecolor{ColorWelsch}{HTML}{911EB4}
\providecolor{ColorWelschmarginal}{HTML}{F032E6}
\providecolor{ColorgauPerPairBest}{HTML}{42D4F4}
\providecolor{Colorigsel}{HTML}{1B7837}
\providecolor{Colormsacsel}{HTML}{D1410A}
\providecolor{Colorlezachlo}{HTML}{F58231}
\providecolor{ColormagsacPerPairBest}{HTML}{F58231}
\providecolor{Colormixlo}{HTML}{911EB4}
\providecolor{ColormsacPerPairBest}{HTML}{FFE119}
\providecolor{Colormsaclo}{HTML}{D4A000}
\providecolor{Colororacleref}{HTML}{BFBFBF}
\providecolor{ColorransacPerPairBest}{HTML}{FFB6C1}
\providecolor{ColorsmoothTLS}{HTML}{469990}
\providecolor{ColorsmoothTLSmarginal}{HTML}{9A6324}
\providecolor{Colorstulo}{HTML}{4363D8}
\providecolor{Colorwelschirlslo}{HTML}{6B8E23}
\providecolor{Colorwelschlo}{HTML}{3CB44B}
\providecolor{dsGeman}{HTML}{E69F00}
\providecolor{dsStut}{HTML}{0072B2}
\providecolor{dsTLS}{HTML}{D62728}
\providecolor{dsWelsch}{HTML}{0B6E4F}
\providecolor{ladderBarronadapt}{HTML}{17BECF}
\providecolor{ladderGNCGM}{HTML}{B8860B}
\providecolor{ladderGNCTLS}{HTML}{C2185B}
\providecolor{ladderHuberECM}{HTML}{E69F00}
\providecolor{ladderHuberfix}{HTML}{E69F00}
\providecolor{ladderL}{HTML}{555555}
\providecolor{ladderLO}{HTML}{1A7A2E}
\providecolor{ladderLift}{HTML}{0B6E4F}
\providecolor{ladderLiftCauchyk}{HTML}{0072B2}
\providecolor{ladderLiftECM}{HTML}{66C2A5}
\providecolor{ladderLiftGMk}{HTML}{7B3294}
\providecolor{ladderLiftN}{HTML}{0B6E4F}
\providecolor{ladderLiftTLSk}{HTML}{E6550D}
\providecolor{ladderLiftk}{HTML}{0B6E4F}
\providecolor{ladderPoseLibLZ}{HTML}{FD8D3C}
\providecolor{ladderPoseLibTLS}{HTML}{E6550D}
\providecolor{ladderReGeMMECM}{HTML}{8C564B}
\providecolor{ladderReGeMMfix}{HTML}{8C564B}
\providecolor{ladderWelschIRLS}{HTML}{6B8E23}
\providecolor{ladderWelschSched}{HTML}{1F77B4}
\providecolor{laddertFullECM}{HTML}{6A3D9A}
\providecolor{laddertFullfix}{HTML}{6A3D9A}
\providecolor{sc0072B2}{HTML}{0072B2}
\providecolor{sc0B6E4F}{HTML}{0B6E4F}
\providecolor{scD62728}{HTML}{D62728}
\providecolor{termGAUsel}{HTML}{42D4F4}
\providecolor{termIGLO}{HTML}{1A7A2E}
\providecolor{termIGLZ}{HTML}{FD8D3C}
\providecolor{termIGLift}{HTML}{0B6E4F}
\providecolor{termIGPoseLib}{HTML}{E6550D}
\providecolor{termIGWIrls}{HTML}{6B8E23}
\providecolor{termIGsel}{HTML}{3CB44B}
\providecolor{termMSACLZ}{HTML}{D62728}
\providecolor{termMSACPoseLib}{HTML}{B5530B}
\providecolor{termMSACsel}{HTML}{F58231}
\providecolor{termoracleref}{HTML}{BFBFBF}
\providecolor{ColorIGTLSHQ}{HTML}{E6550D}
\providecolor{ColorIGGNCW}{HTML}{006D2C}
\providecolor{ColorIGGNCTLS}{HTML}{8C2D04}
\providecolor{ColorIGGMHQ}{HTML}{807DBA}
\providecolor{ColorIGGNCGM}{HTML}{54278F}
\providecolor{ColorIGBCD}{HTML}{08519C}
\providecolor{ColorIGTLSLO}{HTML}{F16913}
\providecolor{ColorIGTLSLOP}{HTML}{A63603}
\providecolor{ColorIGTLSLOPRef}{HTML}{543005}
\providecolor{ColorIGGMIrls}{HTML}{9E9AC8}

\providecolor{Colorigsel}{HTML}{1B7837}
\providecolor{Colormsacsel}{HTML}{D1410A}
\providecolor{Colororacleref}{HTML}{BFBFBF}
\providecolor{Colormsaclo}{HTML}{D4A000}
\providecolor{Colorlezachlo}{HTML}{F58231}
\providecolor{Colorstulo}{HTML}{4363D8}
\providecolor{Colorwelschirlslo}{HTML}{6B8E23}
\providecolor{Colorwelschlo}{HTML}{3CB44B}
\usepackage{mathtools,amssymb}
\usepackage{derivative}
\usepackage{algorithm}
\usepackage{algpseudocode}
\usepackage[accsupp]{axessibility}

\usepackage[hidelinks]{hyperref}
\usepackage[capitalize]{cleveref}

\DeclareMathOperator{\RSS}{RSS}

\DeclareMathOperator*{\argmax}{arg\,max}
\DeclareMathOperator*{\argmin}{arg\,min}

\makeatletter
\@for\@tmp:={A,B,C,D,E,F,G,H,I,J,K,L,M,N,O,P,Q,R,S,T,U,V,W,X,Y,Z,%
              a,b,c,d,e,g,h,i,j,k,l,m,n,o,p,q,r,s,t,u,v,w,x,y,z}\do{%
  \expandafter\edef\csname b\@tmp\endcsname{\noexpand\mathbf{\@tmp}}%
}
\@for\@tmp:={A,B,C,D,E,F,G,H,I,J,K,L,M,N,O,P,Q,R,S,T,U,V,W,X,Y,Z}\do{%
  \expandafter\edef\csname c\@tmp\endcsname{\noexpand\mathcal{\@tmp}}%
}
\@for\@tmp:={alpha,beta,gamma,delta,epsilon,varepsilon,zeta,theta,vartheta,%
              iota,kappa,lambda,mu,nu,xi,pi,varpi,rho,varrho,sigma,varsigma,tau,%
              upsilon,phi,varphi,chi,psi,omega,%
              Gamma,Delta,Theta,Lambda,Xi,Pi,Sigma,Upsilon,Phi,Psi,Omega}\do{%
  \expandafter\edef\csname b\@tmp\endcsname{%
    \noexpand\boldsymbol{\expandafter\noexpand\csname\@tmp\endcsname}%
  }%
}
\makeatother
\newcommand{\bzero}{\mathbf{0}}
\newcommand{\NN}{\mathcal{N}}
\newcommand{\Reals}{\mathbb{R}}
\newcommand{\IG}{\mathrm{IG}}

\newcommand{\stuU}{Stu-$t$U}          
\newcommand{\tlsirls}{TLS-IRLS}   

\newcommand{\kscore}{k_{\mathrm{score}}}
\newcommand{\klo}{k_{\mathrm{LO}}}

\newcommand{\OursCal}{0.66}

\newcommand{\OursInfl}{0.52}
\newcommand{\TLSInfl}{0.07}

\newcommand{\HEBCsMed}{251}
\newcommand{\HEBInlMed}{10}

\newcommand{\PolLiftAfter}{0.662}

\newcommand{\PolBCDAfter}{0.646}
\newcommand{\BasinGnc}{0.13}

\newcommand{\BasinRefinable}{23\,876}   
\newcommand{\BasinHqEight}{0.31}         

\newcommand{\BasinSeedEight}{0.17}

\newcommand{\PTTestPairs}{24\,200}
\newcommand{\AnchorCalDiff}{0.02}
\newcommand{\AnchorInflAnch}{3.1\text{--}3.6}
\newcommand{\AnchorInflFixed}{11\text{--}27}
\newcommand{\BudPTHqLow}{0.54}   
\newcommand{\BudPTNextLow}{0.48}
\newcommand{\BudPTHqHigh}{0.64}  
\newcommand{\BudPTNextHigh}{0.62}
\newcommand{\BudFHq}{0.71}       
\newcommand{\BudFBest}{0.73}
\newcommand{\EtoEOursMed}{1.58}
\newcommand{\EtoESRMed}{2.23}
\newcommand{\EtoEOursMAA}{0.69}
\newcommand{\EtoESRMAA}{0.64}

\newcommand{\EtoEOursRange}{1.8\text{--}6.2}

\newcommand{\EtoESRRangeHi}{54}

\newcommand{\EtoESRInflMed}{38.27}

\newcommand{\suppsec}[1]{\cref{#1}}

\newcommand{\suppfig}[1]{\cref{#1}}
\usepackage{tabularx}
\usepackage{placeins}
\providecolor{Colorigsel}{HTML}{1B7837}
\providecolor{Colormsacsel}{HTML}{D1410A}
\providecolor{Colororacleref}{HTML}{BFBFBF}
\providecolor{Colormsaclo}{HTML}{D4A000}
\providecolor{Colorlezachlo}{HTML}{F58231}
\providecolor{Colorstulo}{HTML}{4363D8}
\providecolor{Colorwelschirlslo}{HTML}{6B8E23}
\providecolor{Colorwelschlo}{HTML}{3CB44B}
\providecolor{ladderBarronadapt}{HTML}{17BECF}
\providecommand{\stuU}{Stu-$t$U}
\providecommand{\kscore}{k_{\mathrm{score}}}
\providecommand{\klo}{k_{\mathrm{LO}}}
\providecommand{\cI}{\mathcal{I}}
\providecommand{\NN}{\mathcal{N}}
\providecommand{\Reals}{\mathbb{R}}
\providecommand{\bzero}{\mathbf{0}}
\providecommand{\be}{\mathbf{e}}
\providecommand{\tlsirls}{TLS-IRLS}

\providecommand{\btheta}{\boldsymbol\theta}

\begin{document}

\title{Robust Local Optimization Done Right}

\titlerunning{Robust Local Optimization Done Right}

\author{James Pritts \and Kevin K\"oser}
\authorrunning{J. Pritts and K\"oser}
\institute{Kiel University, Kiel, Germany}

\maketitle

\begin{abstract}
	RANSAC scoring and local optimization (LO) impose different robustness
	requirements, motivating the separation of hypothesis selection from
	refinement. We systematically isolate the effects of robust-loss shape,
	incorrectly specified inlier scales, and optimization strategy on essential matrix,
	fundamental matrix, and homography estimation. A profile-marginal score
	marginalizes the nuisance inlier scale and selects an inlier partition,
	from which we estimate the scale that sets the LO loss width; this makes
	LO robust to an inlier scale specified too large, whereas one specified too small degrades selection itself.
	Refinement needs gradient from correspondences the seed currently rejects:
	optimizers that reweight from current residuals stay pinned to their seed,
	whereas methods with broad basins recover strongly perturbed seeds yet
	degrade accurate score-selected hypotheses, so basin size alone is
	insufficient to assess RANSAC LO. Joint half-quadratic optimization
	balances the two and is the most consistent strategy across model classes.
	An optimizer matched to the profile-marginal score, which never decreases
	it, does not reach the best accuracy, challenging the prescription that
	scoring and refinement objectives should match. Composed from these
	findings, our RANSAC reduces the median essential-matrix pose error of a
	state-of-the-art RANSAC on PhotoTourism from $2.23^\circ$ to $1.58^\circ$ with a correctly specified inlier scale and
	from $38^\circ$ to $6.2^\circ$ when it is grossly misspecified ($128\times$ too large).\footnote{Project page: \url{https://localoptimizationdoneright.github.io/}.}

\end{abstract}

\section{Introduction}
\label{sec:intro}
Robust geometric estimation from contaminated correspondences is
central to computer vision~\cite{Fischler1981,Huber1981,Maronna2019}.
Modern RANSAC variants separate global search from local optimization
(LO): minimal samples generate hypotheses, a score ranks them, and LO
refines promising hypotheses using their tentative
support~\cite{Chum2003,Lebeda2012}. These stages impose different
requirements: scoring must reliably rank hypotheses, whereas LO needs
informative descent directions despite contamination and an imperfect
initial estimate. Both typically depend on an approximate inlier
scale, whose miscalibration can distort ranking and either amplify
outlier influence or suppress valid measurements during refinement.
Even with a well-calibrated scale, valid measurements may have large
residuals under an imperfect model. Hard truncation, common in RANSAC
scoring functions, provides no gradient beyond its cutoff and can
prevent these measurements from guiding refinement.

Iteratively reweighted least squares
(IRLS)~\cite{HollandWelsch1977,Maronna2019} sets weights from current
residuals, potentially suppressing valid measurements prematurely;
classical LO-RANSAC mitigates this through threshold reduction and
inner resampling~\cite{Chum2003,Lebeda2012}.  Graduated non-convexity
(GNC) instead progressively transforms smoother surrogate losses into
the target
objective~\cite{BlakeZisserman1987,BlackRangarajan1996,YangCarlone2020}.
Joint half-quadratic (HQ) optimization introduces an independent
weight for each observation. Initializing these weights near one lets
large-residual observations continue to influence the model, while
minimizing over the weights recovers the target
objective~\cite{BlackRangarajan1996,Zach2014,ZachBourmaud2018}.

Matching scoring and refinement objectives, with updates accepted
only when they preserve or improve the objective, ensures a
nondecreasing score~\cite{Shekhovtsov2025}, but does not guarantee
geometric accuracy. The scoring objective can also restrict the
optimization strategy, for example by precluding an independent
observation-wise HQ representation. We therefore separate the scoring
rule, scale supply, and optimization path.

Shekhovtsov~\cite{Shekhovtsov2025} unifies Gaussian--uniform mixture
scores through the Bayes decision threshold $\tau$ between inliers and
outliers, which depends on the inlier noise standard deviation
$\sigma$. RANSAC~\cite{Fischler1981}, MSAC and MLESAC~\cite{Torr2000},
and MAGSAC and MAGSAC++~\cite{Barath2019,Barath2020} retain a
residual-unit parameter fixed at test time: a threshold or, for MAGSAC
and MAGSAC++, an upper scale bound $\bar\sigma$. We call these scores
\emph{threshold-based}. SIMFIT~\cite{Heinrich2013} alternates RANSAC
fitting with scale estimation; SIMFIT++~\cite{Edstedt2025} reduces
fitting and truncation bias using held-out residuals and a corrected
estimator. We instead use the profile-marginal score of Pritts
et~al.~\cite{PrittsScoring2026}, which marginalizes the inlier scale
and selects an inlier--outlier partition, retaining only an outlier
half-width in residual units.  It integrates readily into RANSAC, adds
negligible runtime to the RANSAC pipeline, and is the most resilient
to scale miscalibration among the scores evaluated in our
supplementary experiments (\suppsec{sec:supp-scoring}).

We make four contributions. First, we introduce \emph{partition
	anchoring}, which uses the selected inlier partition to estimate the
scale for a separately chosen LO objective, leaving only a dimensionless
width multiplier to tune. Second, controlled comparisons isolate the
effects of scale supply, loss shape, and optimization strategy. Third,
we derive a block-coordinate optimizer, nondecreasing in its score, for the profile-marginal
score and use it as an objective-matched control. Finally, we evaluate
the resulting RANSAC pipeline against
SupeRANSAC~\cite{BarathSupeRANSAC2025} on PhotoTourism and find higher
mAA@$10^\circ$ across the tested scale conditions.

\section{Profile-Marginal Scoring}
\label{sec:scale-free}
Let $\be_i(\btheta)\in\Reals^{d_g}$ denote the residual vector of
observation $i$, where $d_g$ is its dimension, and let
$r_i(\btheta)=\|\be_i(\btheta)\|$ be its norm.
We use the profile-marginal Student-$t$/uniform score of
Pritts et~al.~\cite{PrittsScoring2026}, denoted \stuU{}.
Inlier residuals are Gaussian with scale $\sigma$, while outlier
residuals are uniform on $[-a,a]^{d_g}$. Thus, for an inlier
index set $\cI\subseteq\{1,\ldots,N\}$, the residual-vector
likelihood is
\begin{equation}
	p(\{\be_i\}_{i=1}^N\mid\btheta,\cI,\sigma)
	=\prod_{i\in\cI}\NN\bigl(\be_i(\btheta);\bzero,\sigma^2\bI\bigr)
	\prod_{i\notin\cI}(2a)^{-d_g},
	\label{eq:partition-likelihood}
\end{equation}
where the uniform factors assume that outlier residuals lie
within the stated support.
Let $y_i\in\{0,1\}$ indicate whether observation $i$ is an
inlier, and write $\by=(y_1,\ldots,y_N)$ for the corresponding
labels, so that $\cI=\{i:y_i=1\}$. Let $n_I=|\cI|$ and
$n_O=N-n_I$, and set the prior inlier probability to $\gamma=1/2$.
The joint likelihood of residuals and labels is
\begin{equation}
	p(\{\be_i\}_{i=1}^N,\by\mid\btheta,\sigma)
	=\gamma^{n_I}(1-\gamma)^{n_O}\,
	p(\{\be_i\}_{i=1}^N\mid\btheta,\cI,\sigma),
	\label{eq:joint-partition}
\end{equation}
Its prior factor is constant across partitions.
We drop this common factor from the score.

Rather than marginalizing the labels at a fixed scale, the score
marginalizes the scale for each fixed partition under an
inverse-gamma prior $\pi(\sigma^2)=\IG(\alpha_0,\beta_0)$, with
shape $\alpha_0$ and scale $\beta_0$:
\begin{equation}
	S(\btheta,\cI)=\log\int_0^\infty
	p(\{\be_i\}_{i=1}^N\mid\btheta,\cI,\sigma)\,
	\pi(\sigma^2)\,d\sigma^2.
	\label{eq:score-marginal}
\end{equation}
Jeffreys' prior, $\pi(\sigma^2)\propto1/\sigma^2$, is recovered
as the formal $(\alpha_0,\beta_0)\to(0,0)$ limit, up to a
common normalization constant. We use the degrees-of-freedom adjustment
$|\cI|d_g-n_\theta$ for the $n_\theta=md_g$ fitted model parameters, where
$m$ is the minimal sample size. Under Jeffreys' prior, a partition is
feasible when $|\cI|d_g>n_\theta$, which makes the scale integral finite.
We call maximization over feasible inlier partitions
\emph{partition profiling}; it selects the partition with the highest
scale-marginal score (the $\argmax$ below is over feasible partitions):
\begin{equation}
	\widehat{\cI}(\btheta)\in\argmax_{\cI}S(\btheta,\cI),
	\qquad
	\bar S(\btheta)=S\bigl(\btheta,\widehat{\cI}(\btheta)\bigr).
	\label{eq:profile-score}
\end{equation}
The name \emph{profile-marginal} reflects these two operations:
marginalization over the continuous scale and profiling over
the discrete partition.

The scale integral has a closed form~\cite{PrittsScoring2026}. For a fixed
partition, the score is strictly decreasing in $\RSS_{\cI}(\btheta)$, so
reducing its residual sum of squares raises the score. Under Jeffreys' prior,
the only residual-unit parameter is the outlier half-width $a$.
Our experiments reaffirm the robustness of the profile-marginal score
to extreme scale miscalibration relative to state-of-the-art scores
(\suppsec{sec:supp-scoring}).

Using the same degrees-of-freedom count, the selected partition
supplies the scale estimate
\begin{equation}
	\hat s^2(\btheta)
	=\frac{\RSS_{\widehat{\cI}}(\btheta)+2\beta_0}
	{|\widehat{\cI}|\,d_g-n_\theta+2\alpha_0},
	\qquad \widehat{\cI}=\widehat{\cI}(\btheta),
	\label{eq:partition-scale}
\end{equation}
This is the reciprocal posterior mean precision. Under Jeffreys' prior it
reduces to the residual variance
$\RSS_{\widehat{\cI}}/(|\widehat{\cI}|\,d_g-n_\theta)$
of the selected inliers. This estimate anchors the LO loss width.

\subsection{Objective-matched local optimization}
\label{sec:matching-lo}
We minimize the negative profile-marginal score by block-coordinate
descent (BCD), equivalently maximizing the score. Given $\btheta_t$, the
partition step sweeps feasible prefixes of the sorted residuals: for any
fixed cardinality, the smallest residuals minimize $\RSS_{\cI}$ and
maximize the fixed-partition score. The model block fits the selected
partition by nonlinear least squares, accepting only steps with
$\RSS_{\cI_t}(\btheta_{t+1})\le\RSS_{\cI_t}(\btheta_t)$; the alternating updates are
\begin{equation}
	\cI_t\in\argmax_{\cI} S(\btheta_t,\cI),\qquad
	\btheta_{t+1}\approx\argmin_{\btheta}\RSS_{\cI_t}(\btheta).
	\label{eq:bcd-updates}
\end{equation}
Because the partition step maximizes the score and the least-squares step
does not increase the fixed-partition RSS, the score obeys
\begin{equation}
	\bar S(\btheta_{t+1})
	\ge S(\btheta_{t+1},\cI_t)
	\ge S(\btheta_t,\cI_t)
	=\bar S(\btheta_t).
	\label{eq:bcd-monotonicity}
\end{equation}
We call this method \stuU{}-BCD and
use it as an \emph{objective-matched} control. Its memberships are binary,
but its partition is selected jointly from all residuals rather than by
comparing each residual with a fixed threshold. Its partition step uses
its own outlier half-width, tuned separately (\cref{sec:experiments}), so the
nondecreasing score is the profile-marginal at that half-width, which
need not equal the scoring half-width. The nonseparable score
does not admit the independent observation-wise HQ representation used
below (\cref{sec:lift-theory}).

\section{Robust Losses and Optimization Strategies}
\label{sec:soft-optimization}

We compare IRLS, joint HQ optimization, and GNC within three
robust-loss families: Welsch~\cite{DennisWelsch1978},
Geman--McClure (GM)~\cite{GemanMcClure1985,GemanMcClure1987},
and truncated least squares (TLS). Additional references are
TLS-LZ~\cite{LeZach2021} and classical
TLS-LO$^+$~\cite{Lebeda2012,Sattler2019Github}
(\cref{tab:design,sec:sota}).

For residual norm $r$ and width $c$, the loss functions are
\begin{equation}
	\rho^{(\mathrm W)}=\tfrac{c^2}{2}\bigl(1-e^{-r^2/c^2}\bigr),\qquad
	\rho^{(\mathrm{GM})}=\tfrac{c^2r^2}{2(c^2+r^2)},\qquad
	\rho^{(\mathrm{TLS})}=\tfrac12\min\bigl(r^2,c^2\bigr).
	\label{eq:kernels}
\end{equation}
How each optimizer obtains its width is described in
\cref{sec:scale-interface}.
Welsch and GM retain positive weight at every finite residual;
hard TLS assigns zero weight beyond $c$.
TLS-HQ uses a fixed smooth truncated-quadratic surrogate, whereas
TLS-GNC follows surrogates approaching hard TLS. Thus, comparisons
within the TLS family vary both the objective and optimization path.

\subsection{Iteratively reweighted least squares}

The local objectives are observation-separable:
\begin{equation}
	J(\btheta)=\sum_i \rho\bigl(r_i(\btheta);c\bigr).
	\label{eq:separable}
\end{equation}
IRLS sets $w(r)=\rho'(r)/r$, defined at zero by continuous
extension, and updates the model by weighted nonlinear least squares:
\begin{equation}
	\btheta_{t+1}
	\approx\argmin_{\btheta}
	\sum_i w\!\left(r_i(\btheta_t)\right)r_i(\btheta)^2.
	\label{eq:irls-update}
\end{equation}
We take local solver steps rather than solve each subproblem exactly,
recomputing weights after each model update. Since weights are
prescribed by current residuals, valid measurements can be suppressed
before a poor initial model improves. Solver details are in
\suppsec{sec:supp-lo-details}.

\subsection{Joint half-quadratic optimization}
\label{sec:lift-theory}

Half-quadratic lifting introduces an independent weight variable
per observation~\cite{GemanReynolds1992,BlackRangarajan1996}:
\begin{equation}
	\rho(r;c)
	=\min_z\left[\tfrac12zr^2+\Phi(z;c)\right],
	\qquad z^\star(r)=w(r),
	\label{eq:lift}
\end{equation}
where the regularizer and feasible weight range depend on the loss.
Minimizing over weights recovers the original objective; setting
each weight to its conditional optimum recovers the IRLS weight update.
Following Zach~\cite{Zach2014,ZachBourmaud2018}, we instead initialize
weights near one and optimize them jointly with the model.
The weights can therefore depart from their residual-prescribed
values, allowing initially large residuals to guide refinement.
This permits an implicit broad-to-narrow path under a fixed lifted
objective. Dual regularizers and residualizations are given in
\suppsec{sec:supp-hq-derivations}.

The profile-marginal score couples residuals through the partition RSS
and partition selection, so it does not admit the independent
observation-wise representation in \cref{eq:lift}.
This does not exclude augmentations with coupled variables or
a shared scale.

\subsection{Explicit continuation and classical LO}
\label{sec:gnc}

GNC follows a sequence of surrogate losses, warm-starting each stage
from the preceding solution. For Welsch and GM, we use width
continuation,
\begin{equation}
	F_\mu(\btheta)
	=\sum_i\rho^{(K)}\!\left(r_i(\btheta);\sqrt{\mu}\,c\right),
	\qquad K\in\{\mathrm W,\mathrm{GM}\},
	\label{eq:gnc-family}
\end{equation}
with $\mu$ decreasing to one. TLS-GNC instead uses the surrogate
family of Yang et~al.~\cite{YangCarlone2020}, whose continuous
weight transition approaches hard truncation as $\mu$ increases.
All stages share the same model-update budget; continuation need
not reach the target regime within that budget.
Initialization and schedules are detailed in
\suppsec{sec:supp-lo-details} and \suppsec{sec:supp-gnc}.

Classical LO-RANSAC similarly reduces an enlarged consensus threshold,
but retains binary weights and uses inner RANSAC on non-minimal
consensus samples~\cite{Chum2003,Lebeda2012}.
It therefore combines threshold continuation with stochastic search,
whereas GNC changes the surrogate loss and joint HQ optimizes a
fixed lifted objective.

\subsection{Scale supply to local optimization}
\label{sec:scale-interface}
We distinguish three ways to supply scale. A \emph{partition-anchored}
optimizer takes the partition selected for the seed $\btheta_0$ and sets
\begin{equation}
	\hat s_0^2=\hat s^2(\btheta_0),
	\qquad c=\klo\hat s_0,
	\label{eq:partition-anchor}
\end{equation}
with $\hat s^2$ from \cref{eq:partition-scale}.
The loss width $c$ is fixed during that LO call; only the dimensionless
width multiplier $\klo$ is cross-validated. The width multiplier plays the
role of the tuning constant of robust statistics: in units of the inlier
scale it fixes the loss's efficiency under Gaussian noise, and $c=\klo\hat s_0$
converts it into residual units. In GNC, $c$ is the target
width, while surrogate widths change along the continuation path.
A \emph{fixed-scale} optimizer uses the same absolute width
$c_{\mathrm{ref}}$ for every seed. The profile-marginal optimizer below
is \emph{self-scaling}: its partition and derived scale are updated
from the current model, without a separate LO width.

In the default configurations of \cref{tab:design}, Welsch and
Geman--McClure use partition anchoring for IRLS, HQ, and GNC, as do
TLS-HQ and TLS-GNC; \stuU{}-BCD is self-scaling.
Partition anchoring removes the direct dependence of LO width on the
supplied scale, but the pipeline can still depend on $a$ through the
selected seed and partition.

\begin{table}[t]
	\centering
	\caption{\textbf{Local objectives and optimization strategies.}
		Names follow \emph{objective--strategy}. Entries give
		single-threaded time per LO call in ms on PhotoTourism under
		the shared 25-model-update cap.
		$^{\dagger}$\stuU{}-BCD optimizes the profile-marginal score.
		Timing details are in \suppsec{sec:supp-budget}.}
	\label{tab:design}
	\setlength{\tabcolsep}{3.5pt}
	\small
	\begin{adjustbox}{max width=\linewidth}
\begin{tabular}{lcccc}
	\toprule
	objective/strategy [ms] & IRLS & HQ lift & GNC & other \\
	\midrule
	Welsch & 0.59 & 1.28 & 0.97 & -- \\
	Geman--McClure & 0.55 & 0.78 & 0.72 & -- \\
	Trunc.\ least sq.\ (TLS) & 0.55 & 0.80 & 0.79 & -LZ~0.94, -LO$^+$~0.30 \\
	profile marginal (\stuU{}) & -- & -- & -- & -BCD$^{\dagger}$~1.24 \\
	\bottomrule
\end{tabular}

	\end{adjustbox}
\end{table}

\subsection{State of the art}
\label{sec:sota}
Three established truncated refinements serve as baselines. \tlsirls{} is
PoseLib's truncated-loss Levenberg--Marquardt refinement~\cite{PoseLib},
which recomputes binary weights after every model update, i.e., IRLS
with hard truncation. TLS-LZ~\cite{LeZach2021} approaches the same
truncated objective through a bilevel formulation whose penalty parameter
increases on a fixed schedule. TLS-LO$^+$ is the LO$^+$ procedure of
Lebeda et al.~\cite{Lebeda2012} as implemented in
RansacLib~\cite{Sattler2019Github}: it reduces an enlarged threshold to
its target over iterated least-squares refits, runs an inner RANSAC on
non-minimal samples of the provisional consensus, and keeps a refit only
when it improves the truncated score at the target threshold. All three
are hard-truncation methods with a fixed scale
(\cref{sec:scale-interface}): the same absolute threshold for every seed.

\section{Experiments}
\label{sec:experiments}
We examine four RANSAC design choices: scoring rule, LO objective
(Welsch, GM, TLS, or \stuU{}), scale supply (fixed-scale,
partition-anchored, or self-scaling), and optimization strategy
(IRLS, joint HQ, GNC, BCD, or classical LO). We first measure recovery
from perturbed seeds with fixed loss widths. We then use the profile-marginal
score (\stuU{}; \cref{sec:scale-free}) to evaluate pipelines
under scale corruption and varying hypothesis budgets, with objective-matched
and paired scale-supply controls. ETH3D fundamental-matrix and HEB
homography experiments assess these design choices across model classes.
Every local optimizer has a 25-model-update cap, including all GNC stages.
Full protocols are reported in \suppsec{sec:supp-protocol}.

\begin{table}[!t]
	\centering
	\footnotesize
	\setlength{\tabcolsep}{2pt}
	\caption{\textbf{Evaluation datasets.} Scene and image-pair counts are shown
		for validation and test; there is no separate training split. SP+LG denotes
		SuperPoint+LightGlue.}
	\label{tab:datasets-main}
	\resizebox{\linewidth}{!}{%
		\begin{tabular}{l c l c c}
			\toprule
			Dataset                                  & Transform    & Features     & \shortstack{Validation \\scenes / pairs} &
			\shortstack{Test                                                                                \\scenes / pairs} \\
			\midrule
			PhotoTourism~\cite{Jin2021}              & $\mathbf{E}$ &
			RootSIFT~\cite{Arandjelovic2012}         & 10 / 22\,000 & 11 / 24\,200                          \\
			ETH3D~\cite{Schoeps2017}                 & $\mathbf{F}$ &
			SP+LG~\cite{DeTone2018,Lindenberger2023} &
			6 / 2\,242                               & 7 / 797                                              \\
			HEB~\cite{Barath2023}                    & $\mathbf{H}$ & SIFT         &
			5 / 11\,000                              & 5 / 11\,000                                          \\
			\bottomrule
		\end{tabular}
	}
\end{table}

On PhotoTourism, we use RootSIFT correspondences~\cite{Arandjelovic2012},
capped at $4\,096$ per image pair, and normalize image coordinates by the
image half-diagonal. Scoring and
refinement use Sampson residuals~\cite{Hartley2003}. The reference relative
pose of each pair, the pseudo-ground truth against which errors are measured,
is composed from the scene's bundle-adjusted COLMAP
reconstruction~\cite{Schonberger2016,Jin2021}; its inlier residual variance
$\hat\sigma_p^2$ is the Sampson residual variance of the reconstruction's
track correspondences under that pose. Following
Shekhovtsov~\cite{Shekhovtsov2025}, all methods share the same correspondences
and precomputed minimal essential-matrix hypotheses for each pair.

For HEB homographies, we use four-point hypothesis pools, calibrated Sampson
residuals, and the benchmark's reference pose; the error is the rotation error of the best
homography decomposition, as small baselines leave translation ill-conditioned. For ETH3D
fundamental matrices, we use seven-point hypothesis pools and Sampson
residuals. Intrinsics are used only for evaluation: with $\mathbf{A}_i$ denoting the
image normalization into the canonical frame and $\bar{\mathbf K}_i=\mathbf{A}_i\mathbf
	K_i$, we convert $\mathbf F$ to $\mathbf E=\bar{\mathbf K}_2^\top\mathbf
	F\bar{\mathbf K}_1$.

For relative pose, we evaluate the four essential-matrix decompositions
and report the smallest maximum angular error~\cite{Jin2021}. We define
mAA@$10^\circ$ as
\begin{equation}
	\begin{aligned}
		\mathrm{mAA}@10^\circ
		 & =\frac{1}{10^\circ}\int_0^{10^\circ}
		P(e_{\mathrm{pose}}\le t)\,dt,                                 \\
		\text{s.t.}\quad e_{\mathrm{pose}}
		 & =\min_{j\in\{1,2,3,4\}}\max\bigl(e_R^{(j)},e_t^{(j)}\bigr).
	\end{aligned}
	\label{eq:maa}
\end{equation}
Here $e_R^{(j)}$ and $e_t^{(j)}$ are the rotation and
translation-direction errors of decomposition $j$. In \cref{eq:maa},
$t$ is the pose-error threshold, and $P(e_{\mathrm{pose}}\le t)$ is the
per-scene fraction of pairs whose pose error is at most $t$. We compute mAA
per scene and then average over scenes, with failed estimates contributing zero. For
homographies the error is the pose error of the decomposed
homography, and for fundamental matrices that of the converted essential
matrix. Dataset-specific evaluation details are in
\suppsec{sec:supp-protocol}.

\paragraph{Parameter selection.}
We multiply every supplied residual-unit parameter by $\zeta$: $\zeta=1$ is
a correctly specified inlier scale (\emph{calibration}), and $\zeta<1$ and
$\zeta>1$ specify it too small (\emph{deflation}) and too large
(\emph{inflation}). At calibration, we cross-validate the scoring-function parameter by minimizing
median validation pose error over the pooled validation pairs, using the
selected, unrefined hypotheses. We then tune each optimizer's parameter on
the hypotheses selected by the fixed score, under the same 25-update budget.
Each search uses a log-spaced coarse grid, extended while its minimum lies
on the boundary, followed by a fine grid around that minimum; a minimum that
remains on the boundary is refused rather than used. Selected values are
fixed for testing.
Each dataset has its own validation and test split, and parameters are
re-tuned per dataset, on a stratified quarter of every validation scene's
pairs for PhotoTourism and ETH3D and on every validation pair for HEB.

Each method tunes only its own parameter. The scores tune their
residual-unit parameter: the outlier half-width $a$ for \stuU{} and the
threshold for MSAC. Partition-anchored optimizers tune the width multiplier
$\klo$ and set $c=\klo\hat s_0$ per seed. The fixed-scale baselines,
\tlsirls{}, TLS-LZ, and TLS-LO$^+$, tune their loss width
$c_{\mathrm{ref}}$ directly. \stuU{}-BCD has no loss width; its only
residual-unit parameter is the outlier half-width of its partition
re-selection, which is tuned on the refinement axis. SupeRANSAC's maximum expected noise level is set on the same
validation pairs (\cref{sec:exp-e2e}).

\subsection{Scoring rule}
The scoring-only comparison evaluates PhotoTourism essential-matrix selection
(\cref{tab:datasets-main}): with LO disabled, each score ranks the same
hypothesis pool, and we evaluate its top-ranked model. Scores are tuned on
validation median error at calibration and held fixed as the supplied scale is
corrupted. \suppfig{fig:supp-sel-miscal} reports scene-averaged median pose
error and mAA@$10^\circ$. At calibration, profile-marginal scoring is
nearly tied with SIMFIT++: their median errors are
$2.35^\circ$ and $2.34^\circ$, and their mAA values are $0.622$ and $0.621$,
respectively. At $128\times$ inflation, profile-marginal scoring reaches a
median error of $7.9^\circ$ and $0.451$ mAA, while SIMFIT++ reaches $35^\circ$
and $0.133$ mAA; the remaining scores have median errors of $33^\circ$--
$47^\circ$ and at most $0.07$ mAA. Across the scale-corruption sweep,
profile-marginal scoring is more robust than the alternatives.
This selection-only comparison does not isolate partition selection from
scale estimation; pipeline comparisons therefore use profile-marginal scoring.

\subsection{Recovery from perturbed seeds on real correspondences}
\label{sec:exp-ladder}
To measure recovery under a fixed budget, we bypass hypothesis selection at
test time and perturb the SfM reference pose by $\delta$, applying
equal-magnitude random perturbations to rotation and translation direction.
Every method receives the same seed and original correspondences, including
outliers; only the seed is synthetic.

\begin{figure}[!t]
	\centering
	\begin{tikzpicture}
\begin{groupplot}[
  group style={group size=3 by 1, horizontal sep=6pt, vertical sep=12pt,
    xticklabels at=edge bottom, xlabels at=edge bottom},
  scale only axis, width=0.265\linewidth, height=0.23\linewidth,
  xmode={log}, log basis x={2}, xmin={0.8499857538655519}, xmax={73.60001836514073}, xtick={1.0000,2.0000,4.0000,8.0000,16.0000,32.0000,64.0000}, xticklabels={$1^\circ$,$2^\circ$,$4^\circ$,$8^\circ$,$16^\circ$,$32^\circ$,$64^\circ$}, xmajorgrids, ymajorgrids, grid style={gray!20}, label style={font=\footnotesize}, tick label style={font=\scriptsize}, xlabel={pose perturbation $\delta$},
  title style={font=\footnotesize, align=center}]
\nextgroupplot[ymin={0}, ymax={1}, ytick={0,0.2,0.4,0.6,0.8,1}, minor y tick num={1}, yminorgrids, ylabel={PhotoTourism $\mathbf{E}$\\mAA@$10^\circ$}, ylabel style={align=center}, title={Uniform mixtures\\(TLS, Stu-$t$U)}, legend to name=basinlegend, legend columns=4, legend cell align=left, legend style={draw=none, font=\footnotesize, /tikz/every even column/.append style={column sep=8pt}}]
\addplot[black, thin, dashed, forget plot] coordinates {(1.0000,0.8751) (2.0000,0.7614) (4.0000,0.5514) (8.0000,0.1742) (16.0000,0.0000) (32.0000,0.0000) (64.0000,0.0000)};
\addplot[color=ColorIGPoseLib, mark=o, mark options={solid,fill=ColorIGPoseLib}, line width=0.8pt, mark size=1.6pt, dashed, forget plot] coordinates {(1.0000,0.8309) (2.0000,0.7457) (4.0000,0.5592) (8.0000,0.1950) (16.0000,0.0052) (32.0000,0.0008) (64.0000,0.0001)};
\addplot[color=ColorIGLZ, mark=diamond, mark options={solid,fill=ColorIGLZ}, line width=0.8pt, mark size=1.6pt, dashed, forget plot] coordinates {(1.0000,0.8633) (2.0000,0.7775) (4.0000,0.5821) (8.0000,0.2042) (16.0000,0.0034) (32.0000,0.0002) (64.0000,0.0000)};
\addplot[color=ColorIGTLSHQ, mark=triangle, mark options={solid,fill=ColorIGTLSHQ}, line width=0.8pt, mark size=1.6pt, forget plot] coordinates {(1.0000,0.8008) (2.0000,0.7199) (4.0000,0.5643) (8.0000,0.2525) (16.0000,0.0327) (32.0000,0.0090) (64.0000,0.0033)};
\addplot[color=ColorIGTLSLOP, mark=pentagon*, mark options={solid,fill=ColorIGTLSLOP}, line width=0.8pt, mark size=1.6pt, dashed, forget plot] coordinates {(1.0000,0.8432) (2.0000,0.7389) (4.0000,0.5389) (8.0000,0.1709) (16.0000,0.0000) (32.0000,0.0001) (64.0000,0.0000)};
\addplot[color=ColorIGGNCTLS, mark=diamond, mark options={solid,fill=ColorIGGNCTLS}, line width=0.8pt, mark size=1.6pt, forget plot] coordinates {(1.0000,0.5067) (2.0000,0.5030) (4.0000,0.4625) (8.0000,0.2936) (16.0000,0.1344) (32.0000,0.0648) (64.0000,0.0272)};
\addplot[color=ColorIGBCD, mark=oplus, mark options={solid,fill=ColorIGBCD}, line width=0.8pt, mark size=1.6pt, forget plot] coordinates {(1.0000,0.7938) (2.0000,0.7119) (4.0000,0.5518) (8.0000,0.2228) (16.0000,0.0261) (32.0000,0.0086) (64.0000,0.0040)};
\addlegendimage{black, thin, dashed}\addlegendentry{perturbed seed}
\addlegendimage{color=ColorIGPoseLib, mark=o, mark options={solid,fill=ColorIGPoseLib}, line width=0.8pt, dashed}\addlegendentry{TLS-IRLS}
\addlegendimage{color=ColorIGLZ, mark=diamond, mark options={solid,fill=ColorIGLZ}, line width=0.8pt, dashed}\addlegendentry{TLS-LZ}
\addlegendimage{color=ColorIGTLSHQ, mark=triangle, mark options={solid,fill=ColorIGTLSHQ}, line width=0.8pt}\addlegendentry{TLS-HQ}
\addlegendimage{color=ColorIGTLSLOP, mark=pentagon*, mark options={solid,fill=ColorIGTLSLOP}, line width=0.8pt, dashed}\addlegendentry{TLS-LO$^+$}
\addlegendimage{color=ColorIGGNCTLS, mark=diamond, mark options={solid,fill=ColorIGGNCTLS}, line width=0.8pt}\addlegendentry{TLS-GNC}
\addlegendimage{color=ColorIGBCD, mark=oplus, mark options={solid,fill=ColorIGBCD}, line width=0.8pt}\addlegendentry{Stu-$t$U-BCD}
\addlegendimage{color=ColorIGWIrls, mark=o, mark options={solid,fill=ColorIGWIrls}, line width=0.8pt}\addlegendentry{Welsch-IRLS}
\addlegendimage{color=ColorIGLift, mark=triangle, mark options={solid,fill=ColorIGLift}, line width=0.8pt}\addlegendentry{Welsch-HQ}
\addlegendimage{color=ColorIGGNCW, mark=diamond, mark options={solid,fill=ColorIGGNCW}, line width=0.8pt}\addlegendentry{Welsch-GNC}
\addlegendimage{color=ColorIGGMIrls, mark=o, mark options={solid,fill=ColorIGGMIrls}, line width=0.8pt}\addlegendentry{GM-IRLS}
\addlegendimage{color=ColorIGGMHQ, mark=triangle, mark options={solid,fill=ColorIGGMHQ}, line width=0.8pt}\addlegendentry{GM-HQ}
\addlegendimage{color=ColorIGGNCGM, mark=diamond, mark options={solid,fill=ColorIGGNCGM}, line width=0.8pt}\addlegendentry{GM-GNC}
\nextgroupplot[ymin={0}, ymax={1}, ytick={0,0.2,0.4,0.6,0.8,1}, minor y tick num={1}, yminorgrids, yticklabels={}, title={Welsch}]
\addplot[black, thin, dashed, forget plot] coordinates {(1.0000,0.8751) (2.0000,0.7614) (4.0000,0.5514) (8.0000,0.1742) (16.0000,0.0000) (32.0000,0.0000) (64.0000,0.0000)};
\addplot[color=ColorIGWIrls, mark=o, mark options={solid,fill=ColorIGWIrls}, line width=0.8pt, mark size=1.6pt, forget plot] coordinates {(1.0000,0.7059) (2.0000,0.6190) (4.0000,0.4668) (8.0000,0.1685) (16.0000,0.0081) (32.0000,0.0033) (64.0000,0.0015)};
\addplot[color=ColorIGLift, mark=triangle, mark options={solid,fill=ColorIGLift}, line width=0.8pt, mark size=1.6pt, forget plot] coordinates {(1.0000,0.7654) (2.0000,0.7067) (4.0000,0.5776) (8.0000,0.3144) (16.0000,0.0911) (32.0000,0.0323) (64.0000,0.0114)};
\addplot[color=ColorIGGNCW, mark=diamond, mark options={solid,fill=ColorIGGNCW}, line width=0.8pt, mark size=1.6pt, forget plot] coordinates {(1.0000,0.7338) (2.0000,0.6479) (4.0000,0.4913) (8.0000,0.1781) (16.0000,0.0088) (32.0000,0.0013) (64.0000,0.0001)};
\nextgroupplot[ymin={0}, ymax={1}, ytick={0,0.2,0.4,0.6,0.8,1}, minor y tick num={1}, yminorgrids, yticklabels={}, title={Geman--McClure\\(GM)}]
\addplot[black, thin, dashed, forget plot] coordinates {(1.0000,0.8751) (2.0000,0.7614) (4.0000,0.5514) (8.0000,0.1742) (16.0000,0.0000) (32.0000,0.0000) (64.0000,0.0000)};
\addplot[color=ColorIGGMIrls, mark=o, mark options={solid,fill=ColorIGGMIrls}, line width=0.8pt, mark size=1.6pt, forget plot] coordinates {(1.0000,0.7651) (2.0000,0.6826) (4.0000,0.5154) (8.0000,0.1809) (16.0000,0.0027) (32.0000,0.0003) (64.0000,0.0001)};
\addplot[color=ColorIGGMHQ, mark=triangle, mark options={solid,fill=ColorIGGMHQ}, line width=0.8pt, mark size=1.6pt, forget plot] coordinates {(1.0000,0.7953) (2.0000,0.7218) (4.0000,0.5755) (8.0000,0.2745) (16.0000,0.0532) (32.0000,0.0172) (64.0000,0.0067)};
\addplot[color=ColorIGGNCGM, mark=diamond, mark options={solid,fill=ColorIGGNCGM}, line width=0.8pt, mark size=1.6pt, forget plot] coordinates {(1.0000,0.2576) (2.0000,0.2601) (4.0000,0.2561) (8.0000,0.1516) (16.0000,0.0607) (32.0000,0.0365) (64.0000,0.0176)};
\end{groupplot}
\end{tikzpicture}\\[3pt]
	\pgfplotslegendfromname{basinlegend}
	\caption{\textbf{Recovery from perturbed seeds by loss family.}
		mAA@$10^\circ$ versus pose perturbation $\delta$ on $\BasinRefinable$ of
		$\PTTestPairs$ PhotoTourism test pairs. Each method uses its own
		cross-validated absolute width and the shared 25-model-update budget.
		The dashed grey curve evaluates the unrefined seed. TLS variants
		use the distinct objectives described in \cref{sec:soft-optimization}.
		At $8^\circ$, Welsch-HQ reaches $\BasinHqEight$ mAA@$10^\circ$ versus $\BasinSeedEight$ for
		the unrefined seed; at $16^\circ$, TLS-GNC retains the most accuracy
		($\BasinGnc$ mAA@$10^\circ$).}
	\label{fig:basin-ladder}
\end{figure}
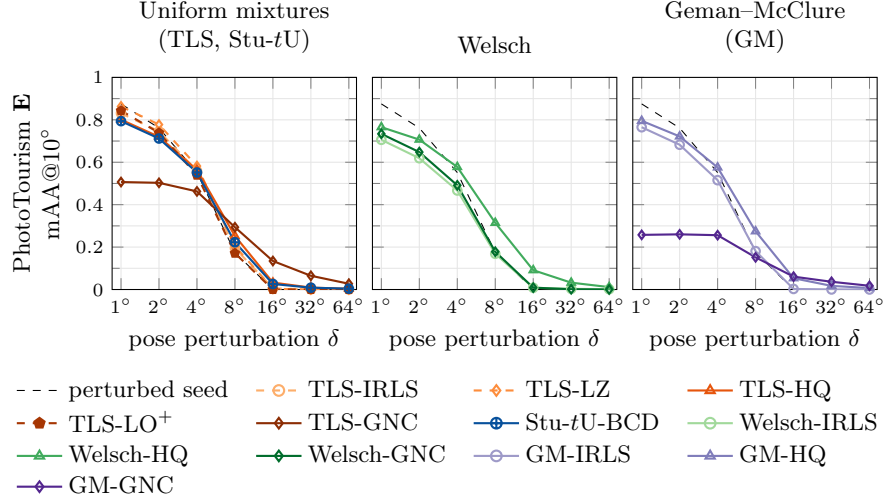

\Cref{fig:basin-ladder} compares IRLS, joint HQ optimization, and GNC for
Welsch and GM, and also includes truncated refits and classical LO for TLS.
The TLS methods can have different terminal objectives
(\cref{sec:soft-optimization}). Because test seeds are supplied directly,
partition-anchored methods tune an absolute width $c$ instead of $\klo$ and
hold it fixed across test pairs and perturbations. Validation otherwise
follows the two-stage protocol in \cref{sec:experiments}; fixed-scale
methods retain their usual tuning. The classical reference is RansacLib's
TLS-LO$^+$~\cite{Sattler2019Github}. Since the reference pose comes from
multi-view SfM while each optimizer fits a two-view pose, even an unperturbed
seed may differ from the reference, affecting scores at small perturbations.

Methods that recompute weights from current residuals stay pinned to the
seed: IRLS, the truncated refits, and \stuU{}-BCD (a binary reweighting,
\cref{eq:bcd-updates}) reach $0.17$--$0.22$ mAA@$10^\circ$ at $8^\circ$
versus $0.17$ for the seed; a monotone score alone does not widen the
basin. Joint HQ instead optimizes the weights with the model
(\cref{eq:lift}), so both move independently, and expands the Welsch and GM
basins. GNC does not help Welsch but recovers the most distant TLS and GM
seeds, at a cost near the seed. Travel thus needs gradient from
correspondences the seed currently rejects, yet that gradient can pull a
good seed toward a contaminated fit; Welsch-HQ balances the two without a
continuation schedule.

\begin{figure}[t!]
	\centering
	\adjustbox{max width=\linewidth}{\input{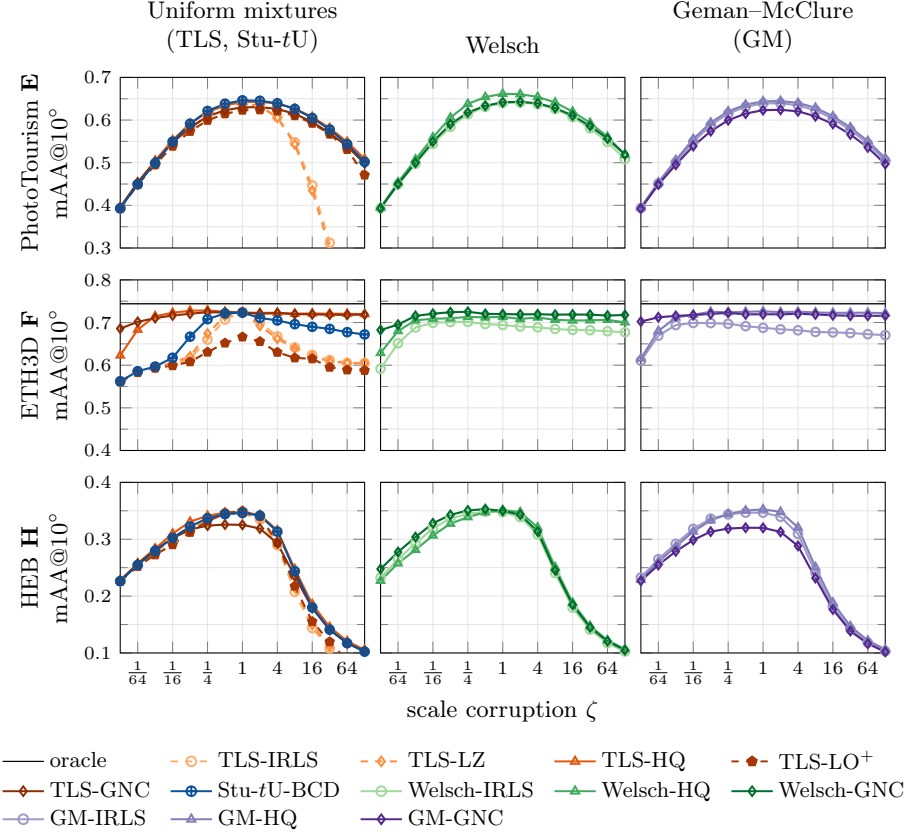}}\\[3pt]
	\adjustbox{max width=\linewidth}{\pgfplotslegendfromname{miscalgridlegend}}
	\caption{\textbf{Composed pipelines under scale miscalibration.}
		Scene-averaged mAA@$10^\circ$ with \stuU{} scoring on PhotoTourism
		essential matrices (top; $\PTTestPairs$ pairs, 11 scenes), ETH3D fundamental
		matrices (middle), and HEB homographies (bottom). Columns group TLS and
		\stuU{}, Welsch, and GM. Parameters are tuned at $\zeta=1$; the
		corruption protocol is described in \cref{sec:exp-miscal-lo}.
		On PhotoTourism essential matrices, Welsch-HQ scores $\OursCal$ mAA@$10^\circ$
		at calibration and retains $\OursInfl$ at $128\times$ inflation, versus $\TLSInfl$
		for fixed-threshold TLS-IRLS. The black reference is oracle selection
		without LO from the minimal-model pool, not a ceiling on refined estimates.}
	\label{fig:miscal}
\end{figure}

\subsection{Composed pipelines under scale miscalibration}
\label{sec:exp-miscal-lo}
We fix profile-marginal scoring (\stuU{}) and compare local optimizers
on the three model classes (\cref{fig:miscal}), tuning each pipeline on
its dataset's validation split (\cref{sec:experiments}). Parameters are
tuned at calibration and held fixed as $\zeta$ varies.
Partition-anchored
methods retain their tuned $\klo$ and infer the scale from the selected
partition. The outlier half-width $a$ remains subject to corruption.

On PhotoTourism, the optimization strategy orders the pipelines more
consistently than the loss. Within each loss, joint HQ is best at
calibration and at nearly every corruption level (Welsch $\OursCal$, GM
$0.64$, TLS $0.64$ mAA@$10^\circ$), while GNC is at most marginally better
than IRLS. Selected seeds usually lie near the solution, where GNC's broadening
is a liability (\cref{sec:exp-ladder}); HQ starts with
weights near one, so every correspondence contributes gradient early, and
tightens the weights under one fixed objective. \stuU{}-BCD, which matches
the scoring objective, is second at $0.65$: objective matching brings it
close, but the composition with a distinct LO objective is better.

Scale corruption splits into two failure modes. Deflation corrupts
selection: at $\zeta=1/16$, all PhotoTourism pipelines lie within $0.02$ of
each other, fixed-scale or not, because the deflated $a$ already degrades
the selected seed and partition, which no optimizer repairs. Inflation
instead corrupts LO: anchored and self-scaling methods retain
$0.50$--$\OursInfl$ at $128\times$, versus $\TLSInfl$ for fixed-threshold
\tlsirls{}. A paired control that changes only scale supply within an
optimizer (\suppsec{sec:supp-anchoring}) confirms this attribution:
anchoring changes pose error by at most $\AnchorCalDiff^\circ$ at $\zeta=1$,
but at $\zeta=128$ reduces it from $\AnchorInflFixed^\circ$ to
$\AnchorInflAnch^\circ$. It does not protect selection: HEB still
degrades under severe inflation.

How much LO can matter depends on the gap to oracle selection. On ETH3D,
the strongest pipelines lie within $0.02$ of the oracle and differ mainly in
robustness to $\zeta$, where HQ and GNC are nearly flat. On HEB, a median of
$\HEBInlMed$ inliers among $\HEBCsMed$ correspondences leaves a gap of about
$0.3$ that no optimizer closes; at calibration the pipelines lie within
$0.01$, except the two GNC variants. There TLS-LO$^+$, the weakest TLS method on
PhotoTourism and ETH3D, ties the strongest; we postulate that its inner
RANSAC on non-minimal samples isolates the small consensus, whereas
reweighting the selected partition is pulled toward structured outliers.
PhotoTourism, with a gap of $0.19$, is where the choice of local optimizer
matters most.

\subsection{Accuracy versus hypothesis budget}
\label{sec:exp-budget}
\begin{figure}[!t]
	\centering
	\adjustbox{max width=\linewidth}{\input{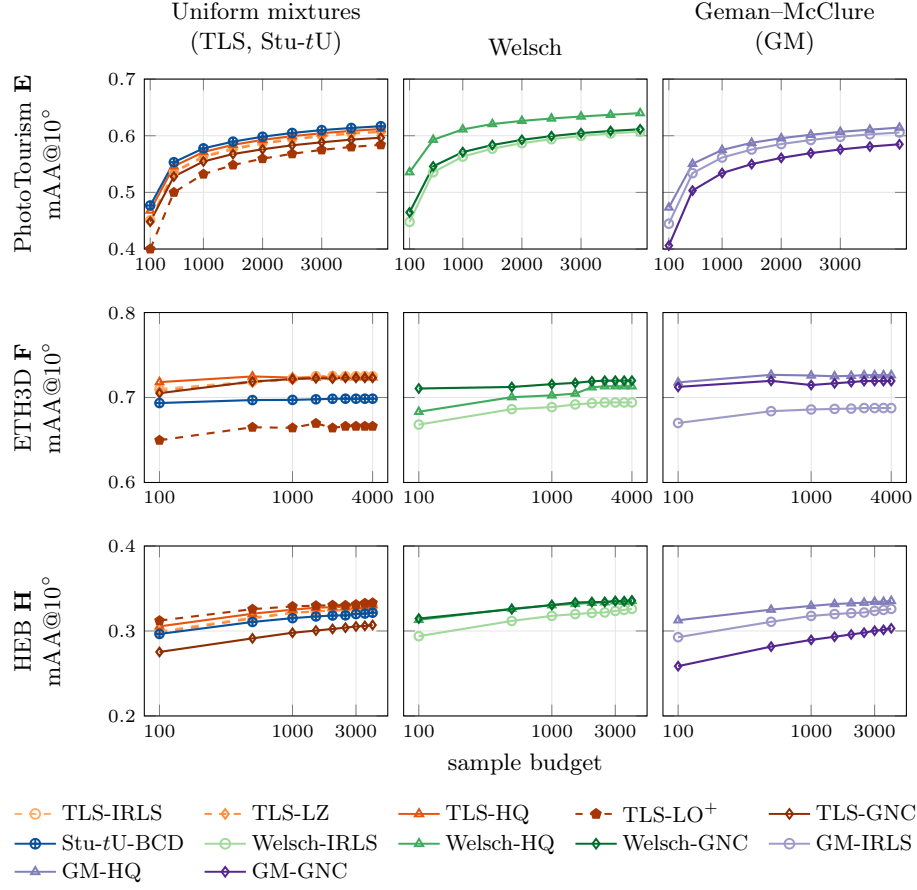}}\\[3pt]
	\adjustbox{max width=\linewidth}{\pgfplotslegendfromname{sbgridlegend}}
	\caption{\textbf{Accuracy versus hypothesis budget.}
		Scene-averaged mAA@$10^\circ$ versus the hypothesis budget, counted in minimal models inspected, on PhotoTourism (top), ETH3D fundamental matrices (middle), and HEB (bottom).
		Each pipeline uses \stuU{} scoring and its dataset-specific
		cross-validated parameters. LO refines the best-so-far hypothesis at
		each checkpoint. On PhotoTourism, Welsch-HQ leads at 100 models
		($\BudPTHqLow$ versus $\BudPTNextLow$ mAA@$10^\circ$ for the next method) and 4,000 models
		($\BudPTHqHigh$ versus $\BudPTNextHigh$). Results are similar across methods on ETH3D and HEB.}
	\label{fig:sample-budget}
\end{figure}
A hypothesis sequence is the ordered list of minimal-sam\-ple models that
RANSAC proposes for an image pair; we draw one per pair and hold it fixed.
The sequence is proposed in the same order to every local optimizer, and
\stuU{} scoring tracks the best hypothesis so far. Local optimization is
evaluated at fixed checkpoints along the sequence, refining the
best-so-far hypothesis at each. This isolates refinement at controlled
hypothesis budgets, with LO triggered only at those checkpoints.
The budget counts minimal models inspected.
\Cref{fig:sample-budget} reports the study on the three model classes, each arm tuned on its dataset's validation split.

\begin{figure}[t]
	\centering
	\input{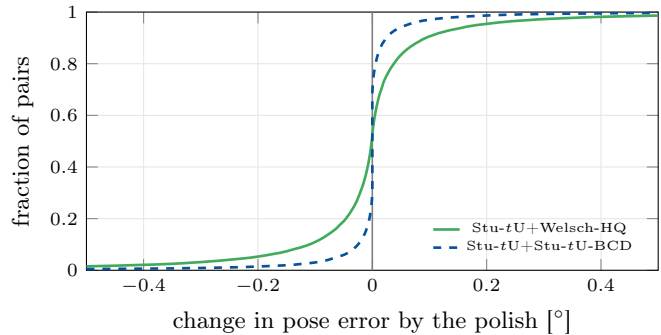}
	\caption{\textbf{Least-squares polish after local optimization.}
	ECDF of the per-pair change in pose error caused by a least-squares
	refit on the inlier partition selected by \stuU{}'s partition-profiling
	step in
	Eq.~\eqref{eq:profile-score} for the LO output,
	$e_{\text{polished}}-e_{\text{LO}}$ (negative is better), on the 11
	PhotoTourism test scenes at $\zeta=1$. Polishing leaves
	mAA@$10^\circ$ unchanged: $\PolLiftAfter$ for Welsch-HQ and $\PolBCDAfter$ for
	\stuU{}-BCD.}
	\label{fig:polish}
\end{figure}

On PhotoTourism, Welsch-HQ is strongest at every budget, and its gain is
largest at small budgets, where selected hypotheses are more likely to lie
outside the basin of IRLS: $\BudPTHqLow$ against $\BudPTNextLow$ for the
next arm after 100 hypotheses, and $\BudPTHqHigh$ against $\BudPTNextHigh$
after 4000. On HEB homographies it is tied with the strongest arms at
every budget. On ETH3D fundamental matrices the arms
lie within 0.04 of each other, except TLS-LO$^+$ (0.67), and Welsch-HQ is not among the
strongest ($\BudFHq$ against $\BudFBest$ for GM-HQ after 4000). Requiring
the optimized model to improve the \stuU{} score does not change the
results. The
timing analysis in \suppfig{fig:supp-budget-time} shows that, beyond the smallest budget, Welsch-HQ retains the best
accuracy at matched LO time despite its higher per-call cost
(\cref{tab:design}).

\subsection{Polishing after local optimization}
\label{sec:exp-polish}
We apply a final nonlinear least-squares refinement to the locally optimized
\stuU{}+Welsch-HQ and \stuU{}+\stuU{}-BCD models on PhotoTourism at the
calibrated scale. For each model, we refit using the inlier partition returned
by \stuU{}'s partition-profiling step in Eq.~\eqref{eq:profile-score}.
As shown in \cref{fig:polish}, pose-error changes are balanced around zero and
the refit yields no systematic pose-accuracy improvement. Starting from the
LO estimate, refining over this fixed inlier set did not measurably
move the solution from the local basin reached by LO in our experiments.

\subsection{RANSAC}
\label{sec:exp-e2e}
The deployed pipeline combines the profile-marginal score of Pritts
et~al.~\cite{PrittsScoring2026} with Welsch-HQ local optimization, whose
loss width is set by partition anchoring
(\cref{eq:profile-score,eq:partition-scale,sec:exp-miscal-lo}). We compare
it with SupeRANSAC~\cite{BarathSupeRANSAC2025}, including sampling and
adaptive termination, on the 11 PhotoTourism test scenes (\cref{tab:e2e}).
Both are tuned at calibration; at test time, $\zeta$ multiplies each
system's residual-unit parameter. At calibration, ours reduces the median
pose error from $\EtoESRMed^\circ$ to $\EtoEOursMed^\circ$ and raises
mAA@$10^\circ$ from $\EtoESRMAA$ to $\EtoEOursMAA$. Under corruption, our
median error stays within $\EtoEOursRange^\circ$, whereas SupeRANSAC's
reaches $\EtoESRRangeHi^\circ$ under deflation and $\EtoESRInflMed^\circ$
at $128\times$ inflation. We postulate that the calibrated gain has the
same origin: a threshold tuned on validation pairs is effectively
miscalibrated for test pairs whose noise scale lies outside the validation
distribution, whereas partition anchoring estimates the scale per pair.

\begin{table}[t]
	\centering
	\small
	\setlength{\tabcolsep}{4pt}
	\caption{\textbf{RANSAC under scale miscalibration.}
		Scene-averaged median pose error, mAA@$10^\circ$ and median wall time
		per pair on the 11 PhotoTourism test scenes, single-threaded; both
		systems tuned at $\zeta=1$, SupeRANSAC at its validated maximum noise level
		with at most $10\,000$ trials. Bold marks the
		better system per column.}
	\label{tab:e2e}
\begin{tabular}{@{}llrrrrr@{}}
 \toprule
 & $\zeta$ & $1/16$ & $1$ & $4$ & $16$ & $128$ \\
 \midrule
 \stuU{}+Welsch-HQ & error [$^\circ$] & \textbf{2.71} & \textbf{1.58} & \textbf{1.84} & \textbf{2.55} & \textbf{6.19} \\
 (Ours) & mAA@$10^\circ$ & \textbf{0.594} & \textbf{0.686} & \textbf{0.666} & \textbf{0.619} & \textbf{0.505} \\
  & time [ms] & 934 & 160 & \textbf{114} & \textbf{74} & \textbf{47} \\
 \midrule
 SupeRANSAC~\cite{BarathSupeRANSAC2025} & error [$^\circ$] & 53.73 & 2.23 & 3.49 & 10.14 & 38.27 \\
  & mAA@$10^\circ$ & 0.210 & 0.643 & 0.558 & 0.333 & 0.072 \\
  & time [ms] & \textbf{92} & \textbf{118} & 176 & 263 & 365 \\
 \bottomrule
\end{tabular}

\end{table}

\section{Conclusion}
\label{sec:discussion}
Our results show that local optimization needs gradient from tentative
correspondences, those that a model fitted to a minimal sample
classifies as outliers, but that the same gradient can pull a good
seed toward a contaminated fit. Methods that reweight from current
residuals, including the objective-matched \stuU{}-BCD, stay pinned to
their seed. Joint HQ lifting balances the two: it is the most
consistent strategy across model classes and performs best on
PhotoTourism, where some score-selected seeds, especially at small
budgets, lie outside the basin of static reweighting. GNC recovers the
most distant seeds but is a liability near the solution, where RANSAC
usually supplies seeds; GM-HQ is among the strongest on ETH3D
fundamental matrices, and LO$^+$ is competitive on the low-inlier HEB
homographies, where its inner RANSAC may isolate a small
consensus. Scale miscalibration splits along the same line: partition
anchoring through the profile-marginal removes the reliance on an
externally calibrated threshold and keeps LO robust to inflation,
whereas deflation degrades selection, which no local optimizer
repairs. Selection also bounds what LO can gain: on HEB, every
optimizer leaves a gap of about $0.3$ to oracle selection. Polishing
by least-squares refitting on selected inliers did not measurably move
the solution out of LO's local basin. An ideal local optimizer could
combine these complementary strategies, motivating further work on
guiding the optimization trajectory beyond the heuristic schedules of
GNC and LO$^+$. Finally, combining the profile-marginal scorer with
partition-anchored, half-quadratically lifted local optimization makes
a state-of-the-art RANSAC pipeline that substantially outperforms a
strong baseline on essential-matrix estimation.

\vspace{\baselineskip}
{\footnotesize\noindent\textbf{Acknowledgements.}~James Pritts is funded by
	Kiel Training for Excellence, an EU Horizon Europe programme under the
	Marie Sk\l{}odowska-Curie Actions (MSCA), grant agreement
	No~101081480. and by Kiel Marine Science
	(KMS).\par}

\bibliographystyle{splncs04}
\bibliography{main}

\clearpage
\appendix
\crefalias{section}{appendix}
\crefalias{subsection}{subappendix}
\crefname{appendix}{Appendix}{Appendices}
\Crefname{appendix}{Appendix}{Appendices}
\crefname{subappendix}{Appendix}{Appendices}
\Crefname{subappendix}{Appendix}{Appendices}
\renewcommand{\topfraction}{0.95}
\renewcommand{\textfraction}{0.05}
\renewcommand{\floatpagefraction}{0.85}
\let\arxivsection\section
\renewcommand{\section}{\FloatBarrier\arxivsection}
\begin{center}\Large\bfseries Supplementary Material\end{center}

\noindent
Notation follows the main paper: $\be_i(\btheta)$ is the residual vector,
$r_i$ its norm, and $\cI$ an inlier partition. This supplement details optimizer implementations
and half-quadratic lifts (\cref{sec:supp-lo-details,sec:supp-hq-derivations}),
additional datasets and protocols (\cref{sec:supp-protocol}), scoring and
scale anchoring under corruption (\cref{sec:supp-scoring,sec:supp-anchoring}),
and the sample-budget runtime analysis (\cref{sec:supp-budget}).

\section{Additional details on local optimization}
\label{sec:supp-lo-details}

\subsection{Solver implementations}

The reference truncated refit, \tlsirls{}, is PoseLib's
truncated-loss Levenberg--Marquardt (LM)
refinement~\cite{PoseLib}, with binary weights recomputed at each
model update. IRLS variants use LM on the model's local
parameterization. The nonlinear weighted least-squares subproblems
are updated locally rather than solved to global optimality.

TLS-HQ uses a fixed smooth truncated-quadratic surrogate whose
weights reach zero at its cutoff. TLS-GNC follows a surrogate
family approaching hard TLS, while TLS-LZ~\cite{LeZach2021}
increases a penalty parameter on a fixed schedule.
These methods belong to the same loss family but need not optimize
identical terminal objectives under the finite update budget.
The classical reference is RansacLib's
TLS-LO$^+$~\cite{Lebeda2012,Sattler2019Github}.

\subsection{IRLS stationarity and HQ lifting}

For a differentiable robust loss with influence function
$\psi=\rho'$, stationarity away from nondifferentiable boundaries
requires
\begin{equation}
	\sum_i\psi(r_i)\nabla r_i
	=\sum_iw(r_i)r_i\nabla r_i=\bzero,
	\qquad w(r)=\psi(r)/r,
	\label{eq:robust-stationarity}
\end{equation}
with $w(0)$ defined by continuous extension.
IRLS alternates these residual-dependent weight assignments with
weighted least-squares model updates.

For the smooth losses considered here, a nonnegative,
non-increasing weight function with finite $w(0)$ makes
$g(u)=\rho(\sqrt{2u};c)$ concave for $u\ge0$, since
$g'(u)=w(\sqrt{2u})$. This yields the multiplicative
HQ representation~\cite{BlackRangarajan1996,ZachBourmaud2018}.
Hard TLS also has an exact lift:
\begin{equation}
	\tfrac12\min(r^2,c^2)
	=\min_{0\le z\le1}
	\left[\tfrac12zr^2+\tfrac12c^2(1-z)\right].
\end{equation}
Our TLS-HQ implementation uses a smooth surrogate for joint
optimization; its dual regularizer and residualization are given
in \cref{sec:supp-hq-derivations}.

Joint HQ updates allow weights to depart temporarily from their
conditional optima. Consequently, neither individual weights nor
true-inlier support must evolve monotonically.
The broad-to-narrow interpretation describes the permitted
optimization path, not a monotonicity constraint.

\subsection{Continuation initialization and limitations}

For Welsch width continuation, we initialize
\begin{equation}
	\mu_0=\max\{1,r_{\max}^2/c^2\},
	\qquad r_{\max}=\max_i r_i(\btheta_0).
\end{equation}
Every initial residual then lies within the initial width
$\sqrt{\mu_0}\,c$. This broadens the region of substantial influence
but does not ensure convexity in the model parameters.

TLS-GNC differs from simply widening a hard cutoff.
Its weights decrease continuously from one to zero across a
transition band; increasing $\mu$ narrows that band and recovers
hard TLS as $\mu\to\infty$. Its schedule therefore runs upward,
whereas the Welsch and GM width schedules decrease toward one.
The TLS-GNC weight is given in closed form in \cref{sec:supp-gnc}.
Under a finite update budget, GNC can follow a contaminated fit
or stop before reaching the target regime.

\subsection{GNC surrogates}
\label{sec:supp-gnc}

For Welsch and GM, GNC uses width continuation: at rung $\mu$, it evaluates
the terminal loss at the inflated width $\sqrt{\mu}\,c$ and decreases $\mu$
toward one. TLS-GNC follows the Black--Rangarajan outer-bound schedule of
Yang et~al.~\cite{YangCarlone2020}, increasing $\mu$ to narrow its continuous
weight-transition band. The 25-model-update budget includes all continuation
stages, so a run may stop before reaching the terminal loss.

For terminal truncation width $\bar c$, the TLS-GNC weight at residual norm
$r$ is
\begin{equation}
w_\mu(r)=
\begin{cases}
1, & r^2\le\dfrac{\mu}{\mu+1}\bar c^2,\\[3pt]
\dfrac{\bar c\sqrt{\mu(\mu+1)}}{|r|}-\mu,
& \dfrac{\mu}{\mu+1}\bar c^2<r^2<\dfrac{\mu+1}{\mu}\bar c^2,\\[6pt]
0, & r^2\ge\dfrac{\mu+1}{\mu}\bar c^2.
\end{cases}
\label{eq:supp-gnc-tls}
\end{equation}
The transition is continuous at both band edges and converges to the hard
truncated least-squares weight as $\mu\to\infty$.

\section{Closed-form half-quadratic lifts}
\label{sec:supp-hq-derivations}

For an even differentiable robust loss with $\rho(0)=0$, multiplicative
half-quadratic lifting writes
\begin{equation}
	\rho(r)
	=
	\min_{z\in\mathcal Z}
	\left[
		\frac12 z r^2+\frac12\kappa^2(z)
		\right].
	\label{eq:supp-hq-form}
\end{equation}
The minimizing quadratic coefficient is the IRLS weight
\begin{equation}
	z^\star=w(r),\qquad
	w(r)=\frac{\rho'(r)}{r}.
	\label{eq:supp-hq-weight}
\end{equation}
At an interior optimum,
\begin{equation}
	\bigl(\kappa^2\bigr)'(z^\star)=-r^2.
\end{equation}
Hence, whenever $w$ is invertible on the relevant branch,
\begin{equation}
	\boxed{
		\bigl(\kappa^2\bigr)'(z)
		=-\bigl(w^{-1}(z)\bigr)^2,
		\qquad
		\kappa^2(z)
		=-\int\bigl(w^{-1}(z)\bigr)^2\,\mathrm dz+C
	}
	\label{eq:supp-hq-master}
\end{equation}
with $C$ chosen so that \cref{eq:supp-hq-form} reproduces $\rho$.
This is the Black--Rangarajan multiplicative construction
\cite{BlackRangarajan1996,Zach2014,ZachBourmaud2018}.

\paragraph{When the lift exists.}
Write $\rho(r)=g(u)$ with $u=r^2/2$, as above. The construction in
\cref{eq:supp-hq-master} is a concave duality in $u$: a regularizer
$\kappa^2$ with the required properties exists exactly when $g$ is
nondecreasing and concave on $u\ge0$ with finite slope at zero,
equivalently when the weight $w(r)=\rho'(r)/r$ is nonincreasing in
$|r|$ with $w(0)<\infty$. This gives a three-tier taxonomy.
\emph{Liftable}: monotone, finite weights; Welsch, Geman--McClure,
Cauchy, Huber, and smooth TLS qualify. \emph{Excluded}: losses with
$w(0)=\infty$, such as $L_1$ and $L_p$ for $p<2$, have an unbounded
line-process domain and are excluded from the bounded multiplicative lift
used here~\cite{BlackRangarajan1996}; losses whose weight increases
anywhere are non-concave in $u$ and admit no multiplicative dual.
\emph{Degenerately liftable}: the exact truncated quadratic
$\rho(r)=\tfrac12\min(r^2,c^2)$ is concave in $u$ and lifts with the
original binary line process, $z\in\{0,1\}$ and
$\kappa^2(z)=c^2(1-z)$, but the $z$-subproblem is linear, so the
optimal weights have no interior and a coupled damped step has no
smooth $z$ direction. Usable joint optimization additionally requires
strict concavity near the cutoff, which is why the HQ arm uses the
smooth TLS kernel below rather than the exact truncation.
Separability is a precondition for the whole framework: the duality
applies one variable per observation to a loss of the form
$\sum_i\rho(r_i)$. Objectives that couple residuals, such as the
profile-marginal score (through the log of the partition's residual
sum of squares and the maximization over partitions) or self-scaling
mixtures (through the jointly estimated scale), admit no
per-observation $g$ and hence no line-process lift; block-coordinate
descent is the direct optimizer of the marginal score.

\paragraph{Welsch.}
For
\begin{equation}
	\rho_{\mathrm W}(r;c)
	=\frac{c^2}{2}\left(1-e^{-r^2/c^2}\right),
\end{equation}
\begin{equation}
	w_{\mathrm W}(r)=e^{-r^2/c^2},
	\qquad
	r^2=-c^2\log z.
\end{equation}
Substitution into \cref{eq:supp-hq-master} and
$\kappa_{\mathrm W}^2(1)=0$ give
\begin{equation}
	\boxed{
		\kappa_{\mathrm W}^2(z)
		=c^2\left(z\log z-z+1\right),
		\qquad 0<z\le1.
	}
	\label{eq:supp-welsch-kappa}
\end{equation}
Therefore
\begin{equation}
	\rho_{\mathrm W}(r;c)
	=
	\min_{0<z\le1}
	\left[
		\frac12zr^2
		+\frac{c^2}{2}\left(z\log z-z+1\right)
		\right].
	\label{eq:supp-welsch-lift}
\end{equation}

\paragraph{Geman--McClure.}
For
\begin{equation}
	\rho_{\mathrm{GM}}(r;c)
	=\frac12\frac{r^2}{1+r^2/c^2},
\end{equation}
\begin{equation}
	w_{\mathrm{GM}}(r)
	=\left(1+r^2/c^2\right)^{-2},
	\qquad
	r^2=c^2\left(z^{-1/2}-1\right).
\end{equation}
Integrating \cref{eq:supp-hq-master} and imposing
$\kappa_{\mathrm{GM}}^2(1)=0$ yield
\begin{equation}
	\boxed{
		\kappa_{\mathrm{GM}}^2(z)
		=c^2\left(1-\sqrt z\right)^2,
		\qquad 0<z\le1,
	}
	\label{eq:supp-gm-kappa}
\end{equation}
and hence
\begin{equation}
	\rho_{\mathrm{GM}}(r;c)
	=
	\min_{0<z\le1}
	\left[
		\frac12zr^2
		+\frac{c^2}{2}\left(1-\sqrt z\right)^2
		\right].
	\label{eq:supp-gm-lift}
\end{equation}

\paragraph{Smooth truncated quadratic.}
We use the normalization
\begin{equation}
	\rho_{\mathrm T}(r;c)
	=
	\begin{cases}
		r^2-\dfrac{r^4}{2c^2}, & |r|\le c, \\[4pt]
		\dfrac{c^2}{2},        & |r|>c.
	\end{cases}
	\label{eq:supp-tls-loss}
\end{equation}
Its quadratic coefficient is
\begin{equation}
	w_{\mathrm T}(r)
	=2\left(1-r^2/c^2\right)_+.
	\label{eq:supp-tls-weight}
\end{equation}
For the interior branch,
$r^2=c^2(1-z/2)$ with $0\le z\le2$. Integrating
\cref{eq:supp-hq-master} and setting $\kappa_{\mathrm T}^2(2)=0$ give
\begin{equation}
	\boxed{
		\kappa_{\mathrm T}^2(z)
		=c^2\left(1-\frac z2\right)^2,
		\qquad 0\le z\le2,
	}
	\label{eq:supp-tls-kappa}
\end{equation}
so
\begin{equation}
	\rho_{\mathrm T}(r;c)
	=
	\min_{0\le z\le2}
	\left[
		\frac12zr^2
		+\frac{c^2}{2}\left(1-\frac z2\right)^2
		\right].
	\label{eq:supp-tls-lift}
\end{equation}
The factor convention in \cref{eq:supp-tls-loss} is chosen to match
\cref{eq:supp-tls-weight,eq:supp-tls-lift}. Multiplying the complete loss by
a positive constant gives an equivalent minimizer but requires scaling both
the quadratic coefficient and the auxiliary penalty consistently.

\paragraph{Least-squares residualization.}
A convenient unconstrained implementation parameterizes the lifted weight as
$z=u^2$ through a square-root confidence $u\in[0,1]$. Each lifted term is then
an ordinary sum of squares:
\begin{align}
	\text{Welsch:}\quad         &
	z=u^2,                      &
	\widehat{\mathbf r}_{\mathrm W}
	                            & =
	\begin{bmatrix}
		ur \\
		c\sqrt{u^2\log u^2-u^2+1}
	\end{bmatrix},
	\label{eq:supp-welsch-residual} \\
	\text{Geman--McClure:}\quad &
	z=u^2,                      &
	\widehat{\mathbf r}_{\mathrm{GM}}
	                            & =
	\begin{bmatrix}
		ur \\
		c(1-u)
	\end{bmatrix},
	\label{eq:supp-gm-residual}     \\
	\text{smooth TLS:}\quad     &
	z=2u^2,                     &
	\widehat{\mathbf r}_{\mathrm T}
	                            & =
	\begin{bmatrix}
		\sqrt2\,ur \\
		c(1-u^2)
	\end{bmatrix}.
	\label{eq:supp-tls-residual}
\end{align}
In each case,
$\rho(r)=\min_u\tfrac12\lVert\widehat{\mathbf r}\rVert^2$.
Initializing $u$ near one gives broad initial support for all three kernels,
despite their different native weight ranges.

All three losses saturate at finite cost. Welsch and Geman--McClure have
$z^\star(r)>0$ for every finite residual and approach zero only
asymptotically; smooth TLS reaches $z^\star=0$ at the finite cutoff
$|r|=c$. Thus a finite auxiliary penalty at zero weight should not be
interpreted as finite-residual rejection for Welsch or Geman--McClure.

\section{Additional datasets and experimental protocol}
\label{sec:supp-protocol}

The main paper reports PhotoTourism essential-matrix experiments with
RootSIFT correspondences~\cite{Jin2021}. The transfer study uses ETH3D
\cite{Schoeps2017} with SuperPoint+LightGlue
\cite{DeTone2018,Lindenberger2023} for fundamental-matrix estimation.

For the fundamental-matrix track, all methods use shared seven-point
hypothesis pools and canonical-frame Sampson residuals. Evaluation converts
$\mathbf F$ to $\mathbf E=\bar{\mathbf K}_2^\top\mathbf F\bar{\mathbf K}_1$,
where $\bar{\mathbf K}_i=\mathbf A_i\mathbf K_i$ uses the image normalization
into the canonical frame, and reports relative-pose error after cheirality
disambiguation. Intrinsics are used only for evaluation.

\paragraph{Controlled comparison.}
The transfer experiments follow the main protocol: all methods use identical
per-pair hypothesis pools, residuals, seeds, and validation splits. For each
pipeline, we tune the scoring parameter first and then tune the optimizer on
the hypotheses selected by that score; all selected parameters are held fixed
on the test scenes. Tuning uses a deterministic stratified quarter of every ETH3D
validation scene's pairs; test scenes are never subsampled.
Local optimizers are capped at 25 model updates;
this is a common nominal cap, not a matched wall-clock budget. Ground truth is
used only for validation, evaluation, and explicitly labeled oracle
experiments.

\paragraph{Classical LO-RANSAC reference.}
We use RansacLib's unmodified TLS-LO$^+$ procedure~\cite{Sattler2019Github}:
the LO trigger, the inner non-minimal
resampling, the threshold schedule, and the least-squares refit hook. RansacLib
ships no two-view solvers, so the model solvers its interface requires were
written for this work: Hartley-normalized eight-point fits with projection to
the essential manifold or to rank two, a four-point fit for homographies, and
squared first-order Sampson residuals. The non-minimal fits are unweighted, as
RansacLib's refit hook drives them; Lebeda et al.~\cite{Lebeda2012} reweighted
their iterated fits by the Sampson gradient.

\paragraph{Metrics.}
Pose error and mAA@$10^\circ$ follow the definitions in the main paper
(Eq.~\ref{eq:maa}). We compute mAA per scene and average over scenes, with
failed estimates contributing zero. The scoring-only comparison reports
scene-averaged median pose error and mAA@$10^\circ$; the basin experiments
report mAA@$10^\circ$.

\paragraph{Scale corruption.}
Methods are tuned once at calibration. At test time, each user-supplied
quantity expressed in residual units is multiplied by
$\zeta\in[1/128,128]$; dimensionless multipliers and scales recomputed from the
selected partition are unchanged. Thus $\zeta$ measures sensitivity to
external calibration rather than perturbing every internal variable
identically. \tlsirls{}-$\hat s$ isolates the effect of replacing a fixed
loss width by one scaled from the selected support.

\section{Scoring under scale miscalibration}
\label{sec:supp-scoring}
This experiment evaluates how well each score selects the seed for LO
under scale miscalibration, with LO disabled: each score ranks the
identical pool of approximately $10^4$ hypotheses per image pair, and the
selected hypothesis is evaluated directly. Cross-validation uses only the
shared scoring window $\kscore\in[1,3.3]$, a dimensionless multiple of
the validation reference scale $s_0$ (\cref{sec:supp-scale-mapping}). At test time
the supplied scale is multiplied by $\zeta\in[1/128,128]$ with the
tuned coordinate $\kscore^*$ held fixed; each score's parameter follows
through its map (\cref{sec:supp-scale-mapping}) and is used for all pairs
without retuning. For \stuU{}, the affected parameter is the outlier
half-width $a$; the selected partition and its scale $\hat s$ remain
data-dependent. SIMFIT++~\cite{Edstedt2025} is MSAC with a data-driven
threshold: starting from an initial threshold, it alternates fitting on a
random half of the correspondences with a bias-corrected median scale
estimate on the other half and scores at the converged threshold; its
initial threshold is cross-validated on the $\tau$ axis like the other
threshold-based scores, and the estimate falls back to it when the
iteration leaves the admissible range. Because that range is the reference's
$0.25$--$8$\,px window, SIMFIT++ scores the pool's residuals re-expressed in
pixels through the stored image normalizations; every other score works in
the canonical frame, where all of its parameters are defined.

\begin{figure}[t]
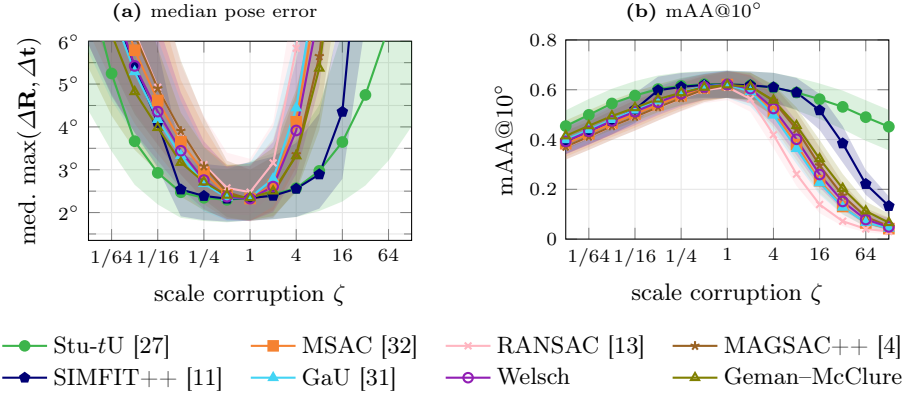

	\centering
	\begin{subfigure}[t]{0.48\linewidth}
		\centering
		\caption{median pose error}
		\adjustbox{max width=\linewidth}{\input{figures/panel_sel_miscal}}
	\end{subfigure}\hfill
	\begin{subfigure}[t]{0.48\linewidth}
		\centering
		\caption{mAA@$10^\circ$}
		\adjustbox{max width=\linewidth}{\input{figures/panel_sel_miscal_maa}}
	\end{subfigure}\\[3pt]
	\adjustbox{max width=\linewidth}{\pgfplotslegendfromname{selmiscallegend}}
	\caption{\textbf{Scoring under scale miscalibration, without LO}
		(PhotoTourism essential matrices; 11 test and 10 validation scenes). We report (a) the
		scene-averaged median pose error, with the axis clipped at $6^\circ$ so
		that the near-tied calibrated regime stays readable (the threshold-based
		scores leave the panel under inflation and reach $33^\circ$--$47^\circ$
		at $128\times$), and (b) mAA@$10^\circ$ at the same operating points,
		which are tuned on the median. The baseline set spans the Welsch and Geman--McClure scores, RANSAC~\cite{Fischler1981},
		MSAC~\cite{Torr1998}, MAGSAC++~\cite{Barath2020}, the
		Gaussian--uniform score~\cite{Shekhovtsov2025}, and SIMFIT++~\cite{Edstedt2025},
		which re-estimates the inlier threshold per pair from held-out
		residuals of the hypothesis selected on a random half of the
		correspondences.}
	\label{fig:supp-sel-miscal}
\end{figure}

The shaded confidence intervals in \cref{fig:supp-sel-miscal} use a
scene-clustered BCa bootstrap~\cite{Efron1987}.

At calibration ($\zeta=1$) the scores are nearly tied
(\cref{fig:supp-sel-miscal}), but \stuU{} is substantially more stable
under scale corruption. This reflects both its data-dependent partition
selection and scale estimation; the half-width map of
\cref{eq:corrupted-cutoff} also shows a logarithmic dependence of the
squared standardized cutoff on $a$, and the experiment
does not isolate these effects. Re-estimating the threshold per pair
(SIMFIT++) does not remove the dependence on the supplied scale: at
$128\times$ inflation its median error is $35^\circ$, inside the
$33^\circ$--$47^\circ$ band of the threshold-based scores, against
$7.9^\circ$ for \stuU{}.

\subsection{Scale corruption and parameter mapping}
\label{sec:supp-scale-mapping}
The shared scoring coordinate $\kscore\in[1,3.3]$ and validation reference
scale $s_0$ determine each score's native residual-unit parameter. RANSAC,
MSAC, SIMFIT++, and the Gaussian--uniform score use a threshold or cutoff
$\tau=\kscore s_0$. MAGSAC++ uses the upper noise-scale bound
$\bar\sigma=\kscore s_0$, not a residual cutoff. For \stuU{}, the outlier
half-width is
\begin{equation}
a(\kscore;s_0,d_g)=s_0\sqrt{\frac{\pi}{2}}
\exp\!\left(\frac{\kscore^2}{2d_g}\right),
\label{eq:native-scale-maps}
\end{equation}
which equates the isotropic Gaussian inlier density and uniform outlier
density at residual norm $\kscore s_0$ under equal assignment priors. This
is a calibration map, not the exact boundary of the profile-marginal
partition. The optimizer uses a separate width coordinate $\klo$.

Replacing $s_0$ by $\zeta s_0$ at fixed $\kscore^*$ multiplies each native
parameter by $\zeta$:
\begin{equation}
\begin{aligned}
\tau(\kscore^*;\zeta s_0)&=\zeta\tau(\kscore^*;s_0),\\
\bar\sigma(\kscore^*;\zeta s_0)&=\zeta\bar\sigma(\kscore^*;s_0),\\
a(\kscore^*;\zeta s_0,d_g)&=\zeta a(\kscore^*;s_0,d_g).
\end{aligned}
\label{eq:native-scale-corruption}
\end{equation}
The scoring-only experiment (\cref{sec:supp-scoring}) applies the nearest
evaluation-grid value to every test pair without retuning. Inverting the maps
at the original reference scale $s_0$ gives the equivalent coordinates
\begin{equation}
k_{\tau}^{\prime}=k_{\bar\sigma}^{\prime}=\zeta\kscore^*,
\qquad
(k_a^{\prime})^2=(\kscore^*)^2+2d_g\log\zeta,
\label{eq:corrupted-cutoff}
\end{equation}
where $k_a^{\prime}$ is real only when the right-hand side is nonnegative.

\section{Paired scale-interface control}
\label{sec:supp-anchoring}

The composed-pipeline study varies both the optimization path and the source
of residual-unit scale. Here we vary only the scale interface within each
optimizer. \tlsirls{}-$\hat s$ replaces a fixed $c_{\mathrm{ref}}$ with
$c=\klo\hat s_0$. For Welsch-GNC and GM-GNC, the paired control replaces
the default partition-derived initial continuation window with one derived
from the externally supplied scale. The terminal loss, optimizer, and validation
protocol are otherwise unchanged.

\begin{figure}[t]
\centering
\adjustbox{max width=\linewidth}{\begin{tikzpicture}
\begin{groupplot}[
  group style={group size=2 by 1, horizontal sep=34pt},
  scale only axis, width=0.40\linewidth, height=0.28\linewidth,
  xmode={log}, log basis x={2}, xlabel={scale corruption $\zeta$}, xmajorgrids, ymajorgrids, grid style={gray!20}, label style={font=\footnotesize}, tick label style={font=\scriptsize}, xmin={0.0078125}, xmax={128.0}, xtick={0.015625,0.0625,0.25,1.0,4.0,16.0,64.0}, xticklabels={$1/64$,$1/16$,$1/4$,$1$,$4$,$16$,$64$}]
\nextgroupplot[ymin={0.4}, ymax={6}, ytick={1,2,3,4,5,6}, minor y tick num={1}, yminorgrids, scaled y ticks={false}, yticklabel style={/pgf/number format/.cd, fixed, precision=0}, yticklabel={$\pgfmathprintnumber{\tick}^\circ$}, ylabel={med.\ $\max(\Delta \mathbf{R},\Delta \mathbf{t})$}, legend to name=anchorlegend, legend columns=3, legend cell align=left, legend style={draw=none, font=\footnotesize, /tikz/every even column/.append style={column sep=8pt}}]
\addplot[color=ColorIGPoseLib, mark=o, mark options={solid,fill=ColorIGPoseLib}, line width=0.8pt, mark size=1.6pt, dashed, forget plot] coordinates {(0.007812,6.8404) (0.015625,4.6256) (0.031250,3.4068) (0.062500,2.6323) (0.125000,2.2424) (0.250000,1.9983) (0.500000,1.8941) (1.000000,1.8712) (2.000000,1.9190) (4.000000,2.0672) (8.000000,2.4977) (16.000000,3.6919) (32.000000,6.9524) (64.000000,13.4370) (128.000000,19.8414)};
\addplot[color=ColorIGPoseLib, mark=o, mark options={solid,fill=ColorIGPoseLib}, line width=0.8pt, mark size=1.6pt, forget plot] coordinates {(0.007812,6.8139) (0.015625,4.5958) (0.031250,3.3752) (0.062500,2.6233) (0.125000,2.2383) (0.250000,2.0014) (0.500000,1.8945) (1.000000,1.8570) (2.000000,1.8729) (4.000000,1.9104) (8.000000,2.0267) (16.000000,2.1949) (32.000000,2.4203) (64.000000,2.7917) (128.000000,3.4624)};
\addplot[color=ColorIGGNCW, mark=diamond, mark options={solid,fill=ColorIGGNCW}, line width=0.8pt, mark size=1.6pt, dashed, forget plot] coordinates {(0.007812,6.9824) (0.015625,4.8543) (0.031250,3.6938) (0.062500,2.8673) (0.125000,2.3164) (0.250000,1.9719) (0.500000,1.8157) (1.000000,1.7950) (2.000000,1.8421) (4.000000,1.9267) (8.000000,2.1007) (16.000000,2.5357) (32.000000,3.6254) (64.000000,6.4636) (128.000000,11.4238)};
\addplot[color=ColorIGGNCW, mark=diamond, mark options={solid,fill=ColorIGGNCW}, line width=0.8pt, mark size=1.6pt, forget plot] coordinates {(0.007812,6.9854) (0.015625,4.8297) (0.031250,3.6191) (0.062500,2.7301) (0.125000,2.2042) (0.250000,1.9363) (0.500000,1.8165) (1.000000,1.7785) (2.000000,1.7944) (4.000000,1.8377) (8.000000,1.9150) (16.000000,2.0659) (32.000000,2.2456) (64.000000,2.5591) (128.000000,3.1232)};
\addplot[color=ColorIGGNCGM, mark=diamond, mark options={solid,fill=ColorIGGNCGM}, line width=0.8pt, mark size=1.6pt, dashed, forget plot] coordinates {(0.007812,7.0095) (0.015625,4.8661) (0.031250,3.7319) (0.062500,2.9349) (0.125000,2.4874) (0.250000,2.2246) (0.500000,2.1167) (1.000000,2.0724) (2.000000,2.0933) (4.000000,2.2712) (8.000000,3.1497) (16.000000,6.4014) (32.000000,11.4482) (64.000000,18.0006) (128.000000,26.7278)};
\addplot[color=ColorIGGNCGM, mark=diamond, mark options={solid,fill=ColorIGGNCGM}, line width=0.8pt, mark size=1.6pt, forget plot] coordinates {(0.007812,7.0073) (0.015625,4.8658) (0.031250,3.7315) (0.062500,2.9306) (0.125000,2.4869) (0.250000,2.2241) (0.500000,2.1156) (1.000000,2.0712) (2.000000,2.0676) (4.000000,2.1013) (8.000000,2.1941) (16.000000,2.3503) (32.000000,2.5665) (64.000000,2.9214) (128.000000,3.5967)};
\addplot[color=ColorIGLift, mark=triangle, mark options={solid,fill=ColorIGLift}, line width=0.6pt, mark size=1.6pt, forget plot] coordinates {(0.007812,6.9198) (0.015625,4.7119) (0.031250,3.4504) (0.062500,2.5268) (0.125000,1.9419) (0.250000,1.6579) (0.500000,1.5654) (1.000000,1.5525) (2.000000,1.5765) (4.000000,1.6323) (8.000000,1.7288) (16.000000,1.8770) (32.000000,2.0926) (64.000000,2.4287) (128.000000,3.0659)};
\addlegendimage{color=ColorIGPoseLib, mark=o, mark options={solid,fill=ColorIGPoseLib}, line width=0.8pt, mark size=1.6pt, dashed}\addlegendentry{TLS-IRLS (fixed scale)}
\addlegendimage{color=ColorIGPoseLib, mark=o, mark options={solid,fill=ColorIGPoseLib}, line width=0.8pt, mark size=1.6pt}\addlegendentry{TLS-IRLS-$\hat s$}
\addlegendimage{color=ColorIGGNCW, mark=diamond, mark options={solid,fill=ColorIGGNCW}, line width=0.8pt, mark size=1.6pt, dashed}\addlegendentry{Welsch-GNC (fixed scale)}
\addlegendimage{color=ColorIGGNCW, mark=diamond, mark options={solid,fill=ColorIGGNCW}, line width=0.8pt, mark size=1.6pt}\addlegendentry{Welsch-GNC}
\addlegendimage{color=ColorIGGNCGM, mark=diamond, mark options={solid,fill=ColorIGGNCGM}, line width=0.8pt, mark size=1.6pt, dashed}\addlegendentry{GM-GNC (fixed scale)}
\addlegendimage{color=ColorIGGNCGM, mark=diamond, mark options={solid,fill=ColorIGGNCGM}, line width=0.8pt, mark size=1.6pt}\addlegendentry{GM-GNC}
\addlegendimage{color=ColorIGLift, mark=triangle, mark options={solid,fill=ColorIGLift}, line width=0.6pt, mark size=1.6pt}\addlegendentry{Welsch-HQ (ref.)}
\nextgroupplot[ymin={0.2}, ymax={1}, ytick={0.2,0.4,0.6,0.8,1}, minor y tick num={1}, yminorgrids, ylabel={mAA@$10^\circ$}]
\addplot[color=ColorIGPoseLib, mark=o, mark options={solid,fill=ColorIGPoseLib}, line width=0.8pt, mark size=1.6pt, dashed, forget plot] coordinates {(0.007812,0.4023) (0.015625,0.4618) (0.031250,0.5111) (0.062500,0.5546) (0.125000,0.5885) (0.250000,0.6135) (0.500000,0.6283) (1.000000,0.6353) (2.000000,0.6333) (4.000000,0.6204) (8.000000,0.5836) (16.000000,0.5050) (32.000000,0.3801) (64.000000,0.2434) (128.000000,0.1567)};
\addplot[color=ColorIGPoseLib, mark=o, mark options={solid,fill=ColorIGPoseLib}, line width=0.8pt, mark size=1.6pt, forget plot] coordinates {(0.007812,0.4038) (0.015625,0.4633) (0.031250,0.5122) (0.062500,0.5552) (0.125000,0.5888) (0.250000,0.6136) (0.500000,0.6282) (1.000000,0.6359) (2.000000,0.6359) (4.000000,0.6303) (8.000000,0.6178) (16.000000,0.5979) (32.000000,0.5722) (64.000000,0.5407) (128.000000,0.5015)};
\addplot[color=ColorIGGNCW, mark=diamond, mark options={solid,fill=ColorIGGNCW}, line width=0.8pt, mark size=1.6pt, dashed, forget plot] coordinates {(0.007812,0.3928) (0.015625,0.4490) (0.031250,0.4965) (0.062500,0.5423) (0.125000,0.5831) (0.250000,0.6153) (0.500000,0.6333) (1.000000,0.6405) (2.000000,0.6395) (4.000000,0.6333) (8.000000,0.6165) (16.000000,0.5799) (32.000000,0.5118) (64.000000,0.4016) (128.000000,0.2864)};
\addplot[color=ColorIGGNCW, mark=diamond, mark options={solid,fill=ColorIGGNCW}, line width=0.8pt, mark size=1.6pt, forget plot] coordinates {(0.007812,0.3930) (0.015625,0.4499) (0.031250,0.4999) (0.062500,0.5494) (0.125000,0.5902) (0.250000,0.6177) (0.500000,0.6335) (1.000000,0.6417) (2.000000,0.6429) (4.000000,0.6391) (8.000000,0.6284) (16.000000,0.6105) (32.000000,0.5862) (64.000000,0.5563) (128.000000,0.5190)};
\addplot[color=ColorIGGNCGM, mark=diamond, mark options={solid,fill=ColorIGGNCGM}, line width=0.8pt, mark size=1.6pt, dashed, forget plot] coordinates {(0.007812,0.3924) (0.015625,0.4486) (0.031250,0.4952) (0.062500,0.5388) (0.125000,0.5735) (0.250000,0.5993) (0.500000,0.6152) (1.000000,0.6231) (2.000000,0.6218) (4.000000,0.6038) (8.000000,0.5347) (16.000000,0.4134) (32.000000,0.2882) (64.000000,0.1778) (128.000000,0.0929)};
\addplot[color=ColorIGGNCGM, mark=diamond, mark options={solid,fill=ColorIGGNCGM}, line width=0.8pt, mark size=1.6pt, forget plot] coordinates {(0.007812,0.3926) (0.015625,0.4487) (0.031250,0.4952) (0.062500,0.5389) (0.125000,0.5738) (0.250000,0.5996) (0.500000,0.6152) (1.000000,0.6232) (2.000000,0.6240) (4.000000,0.6205) (8.000000,0.6088) (16.000000,0.5909) (32.000000,0.5662) (64.000000,0.5357) (128.000000,0.4967)};
\addplot[color=ColorIGLift, mark=triangle, mark options={solid,fill=ColorIGLift}, line width=0.6pt, mark size=1.6pt, forget plot] coordinates {(0.007812,0.3951) (0.015625,0.4539) (0.031250,0.5074) (0.062500,0.5594) (0.125000,0.6057) (0.250000,0.6378) (0.500000,0.6539) (1.000000,0.6615) (2.000000,0.6606) (4.000000,0.6543) (8.000000,0.6413) (16.000000,0.6199) (32.000000,0.5935) (64.000000,0.5607) (128.000000,0.5190)};
\end{groupplot}
\end{tikzpicture}}\\[3pt]
\adjustbox{max width=\linewidth}{\pgfplotslegendfromname{anchorlegend}}
\caption{\textbf{Partition anchoring, paired within optimizer.}
Scene-averaged median pose error and mAA@$10^\circ$ on the 11 PhotoTourism test scenes.
Dashed curves use a fixed reference threshold or a scale-corrupted GNC
window; solid curves use the same optimizer with partition-derived scale.
At $\zeta=128$, anchored variants reach $3.1$--$3.6^\circ$, versus
$11$--$27^\circ$ for their fixed-scale counterparts. The vertical axis is
clipped at $6^\circ$.}
\label{fig:supp-anchoring}
\end{figure}
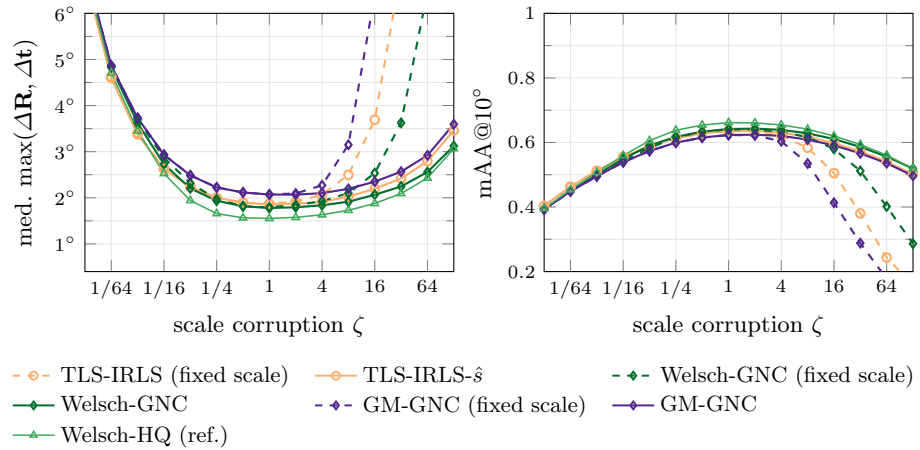

At calibration, pairing changes median pose error by at most $0.02^\circ$.
Under inflation, partition anchoring prevents the large error increase seen
with externally supplied scale. Scale supply therefore explains most of the
inflation sensitivity, but not the basin advantage of Welsch-HQ, which remains
when the scale interface is controlled.

\section{Sample budget and runtime}
	\label{sec:supp-budget}

	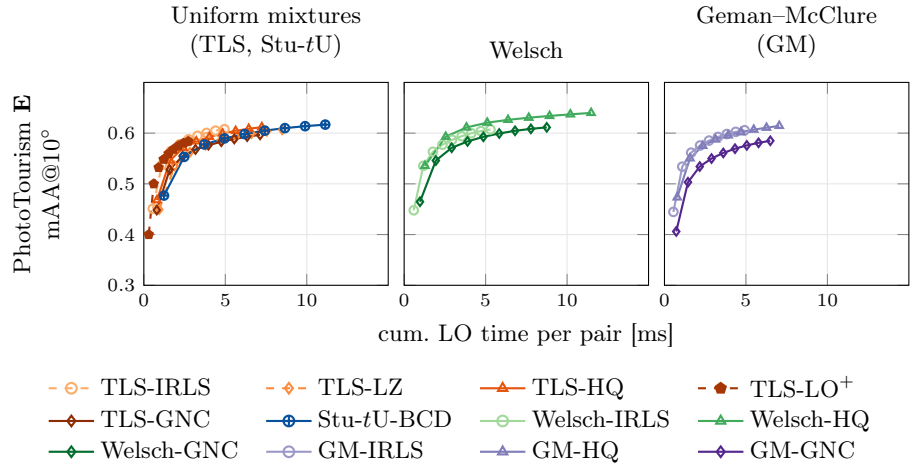
\begin{figure}[t]
		\centering
		\adjustbox{max width=\linewidth}{\begin{tikzpicture}
\begin{groupplot}[
  group style={group size=3 by 1, horizontal sep=6pt, vertical sep=12pt,
    xticklabels at=edge bottom, xlabels at=edge bottom},
  scale only axis, width=0.265\linewidth, height=0.22\linewidth,
  xmin={0}, xmax={15}, xtick={0,5,10},
  xmajorgrids, ymajorgrids, grid style={gray!20},
  label style={font=\footnotesize}, tick label style={font=\scriptsize},
  title style={font=\footnotesize, align=center}]
\nextgroupplot[ymin={0.30}, ymax={0.70}, ytick={0.3,0.4,0.5,0.6}, minor y tick num={0}, yminorgrids, yticklabel style={text width=2.4em, align=right}, title={Uniform mixtures\\(TLS, Stu-$t$U)}, ylabel={PhotoTourism $\mathbf{E}$\\mAA@$10^\circ$}, ylabel style={align=center}, legend to name=sbtimelegend, legend columns=4, legend cell align=left, legend style={draw=none, font=\footnotesize, /tikz/every even column/.append style={column sep=8pt}}]
\addplot[color=ColorIGPoseLib, mark=o, mark options={solid,fill=ColorIGPoseLib}, line width=0.8pt, mark size=1.6pt, dashed, forget plot] coordinates {(0.550,0.4509) (1.099,0.5369) (1.649,0.5642) (2.198,0.5779) (2.748,0.5877) (3.297,0.5947) (3.847,0.6004) (4.396,0.6049) (4.946,0.6078)};
\addplot[color=ColorIGLZ, mark=diamond, mark options={solid,fill=ColorIGLZ}, line width=0.8pt, mark size=1.6pt, dashed, forget plot] coordinates {(0.944,0.4493) (1.888,0.5352) (2.832,0.5625) (3.776,0.5761) (4.719,0.5860) (5.663,0.5933) (6.607,0.5992) (7.551,0.6036) (8.495,0.6066)};
\addplot[color=ColorIGTLSHQ, mark=triangle, mark options={solid,fill=ColorIGTLSHQ}, line width=0.8pt, mark size=1.6pt, forget plot] coordinates {(0.805,0.4679) (1.610,0.5460) (2.414,0.5712) (3.219,0.5839) (4.024,0.5927) (4.829,0.5992) (5.633,0.6044) (6.438,0.6086) (7.243,0.6118)};
\addplot[color=ColorIGTLSLOP, mark=pentagon*, mark options={solid,fill=ColorIGTLSLOP}, line width=0.8pt, mark size=1.6pt, dashed, forget plot] coordinates {(0.303,0.4000) (0.607,0.5000) (0.910,0.5322) (1.213,0.5483) (1.517,0.5595) (1.820,0.5681) (2.123,0.5750) (2.427,0.5802) (2.730,0.5839)};
\addplot[color=ColorIGGNCTLS, mark=diamond, mark options={solid,fill=ColorIGGNCTLS}, line width=0.8pt, mark size=1.6pt, forget plot] coordinates {(0.793,0.4483) (1.586,0.5280) (2.378,0.5546) (3.171,0.5677) (3.964,0.5761) (4.757,0.5831) (5.549,0.5884) (6.342,0.5931) (7.135,0.5965)};
\addplot[color=ColorIGBCD, mark=oplus, mark options={solid,fill=ColorIGBCD}, line width=0.8pt, mark size=1.6pt, forget plot] coordinates {(1.237,0.4769) (2.474,0.5532) (3.712,0.5777) (4.949,0.5896) (6.186,0.5982) (7.423,0.6049) (8.661,0.6099) (9.898,0.6137) (11.135,0.6168)};
\addlegendimage{color=ColorIGPoseLib, mark=o, mark options={solid,fill=ColorIGPoseLib}, line width=0.8pt, dashed}\addlegendentry{TLS-IRLS}
\addlegendimage{color=ColorIGLZ, mark=diamond, mark options={solid,fill=ColorIGLZ}, line width=0.8pt, dashed}\addlegendentry{TLS-LZ}
\addlegendimage{color=ColorIGTLSHQ, mark=triangle, mark options={solid,fill=ColorIGTLSHQ}, line width=0.8pt}\addlegendentry{TLS-HQ}
\addlegendimage{color=ColorIGTLSLOP, mark=pentagon*, mark options={solid,fill=ColorIGTLSLOP}, line width=0.8pt, dashed}\addlegendentry{TLS-LO$^+$}
\addlegendimage{color=ColorIGGNCTLS, mark=diamond, mark options={solid,fill=ColorIGGNCTLS}, line width=0.8pt}\addlegendentry{TLS-GNC}
\addlegendimage{color=ColorIGBCD, mark=oplus, mark options={solid,fill=ColorIGBCD}, line width=0.8pt}\addlegendentry{Stu-$t$U-BCD}
\addlegendimage{color=ColorIGWIrls, mark=o, mark options={solid,fill=ColorIGWIrls}, line width=0.8pt}\addlegendentry{Welsch-IRLS}
\addlegendimage{color=ColorIGLift, mark=triangle, mark options={solid,fill=ColorIGLift}, line width=0.8pt}\addlegendentry{Welsch-HQ}
\addlegendimage{color=ColorIGGNCW, mark=diamond, mark options={solid,fill=ColorIGGNCW}, line width=0.8pt}\addlegendentry{Welsch-GNC}
\addlegendimage{color=ColorIGGMIrls, mark=o, mark options={solid,fill=ColorIGGMIrls}, line width=0.8pt}\addlegendentry{GM-IRLS}
\addlegendimage{color=ColorIGGMHQ, mark=triangle, mark options={solid,fill=ColorIGGMHQ}, line width=0.8pt}\addlegendentry{GM-HQ}
\addlegendimage{color=ColorIGGNCGM, mark=diamond, mark options={solid,fill=ColorIGGNCGM}, line width=0.8pt}\addlegendentry{GM-GNC}
\nextgroupplot[ymin={0.30}, ymax={0.70}, ytick={0.3,0.4,0.5,0.6}, minor y tick num={0}, yminorgrids, yticklabel style={text width=2.4em, align=right}, title={Welsch}, yticklabels={}, xlabel={cum.\ LO time per pair [ms]}]
\addplot[color=ColorIGWIrls, mark=o, mark options={solid,fill=ColorIGWIrls}, line width=0.8pt, mark size=1.6pt, forget plot] coordinates {(0.589,0.4479) (1.178,0.5356) (1.766,0.5632) (2.355,0.5772) (2.944,0.5869) (3.533,0.5939) (4.122,0.5997) (4.711,0.6040) (5.299,0.6072)};
\addplot[color=ColorIGLift, mark=triangle, mark options={solid,fill=ColorIGLift}, line width=0.8pt, mark size=1.6pt, forget plot] coordinates {(1.275,0.5356) (2.550,0.5929) (3.825,0.6112) (5.100,0.6206) (6.375,0.6262) (7.650,0.6304) (8.925,0.6339) (10.200,0.6370) (11.475,0.6399)};
\addplot[color=ColorIGGNCW, mark=diamond, mark options={solid,fill=ColorIGGNCW}, line width=0.8pt, mark size=1.6pt, forget plot] coordinates {(0.972,0.4647) (1.945,0.5459) (2.917,0.5711) (3.890,0.5836) (4.862,0.5927) (5.834,0.5993) (6.807,0.6046) (7.779,0.6084) (8.752,0.6114)};
\nextgroupplot[ymin={0.30}, ymax={0.70}, ytick={0.3,0.4,0.5,0.6}, minor y tick num={0}, yminorgrids, yticklabel style={text width=2.4em, align=right}, title={Geman--McClure\\(GM)}, yticklabels={}]
\addplot[color=ColorIGGMIrls, mark=o, mark options={solid,fill=ColorIGGMIrls}, line width=0.8pt, mark size=1.6pt, forget plot] coordinates {(0.547,0.4447) (1.093,0.5339) (1.640,0.5615) (2.186,0.5756) (2.733,0.5853) (3.280,0.5925) (3.826,0.5982) (4.373,0.6025) (4.920,0.6057)};
\addplot[color=ColorIGGMHQ, mark=triangle, mark options={solid,fill=ColorIGGMHQ}, line width=0.8pt, mark size=1.6pt, forget plot] coordinates {(0.783,0.4733) (1.565,0.5503) (2.348,0.5749) (3.131,0.5875) (3.913,0.5956) (4.696,0.6017) (5.479,0.6068) (6.262,0.6107) (7.044,0.6143)};
\addplot[color=ColorIGGNCGM, mark=diamond, mark options={solid,fill=ColorIGGNCGM}, line width=0.8pt, mark size=1.6pt, forget plot] coordinates {(0.722,0.4061) (1.443,0.5032) (2.165,0.5341) (2.887,0.5499) (3.609,0.5609) (4.330,0.5693) (5.052,0.5758) (5.774,0.5809) (6.496,0.5849)};
\end{groupplot}
\end{tikzpicture}}\\[3pt]
		\adjustbox{max width=\linewidth}{\pgfplotslegendfromname{sbtimelegend}}
		\caption{\textbf{Accuracy versus cumulative LO time on PhotoTourism.}
			The sample-budget checkpoints from the main paper are re-plotted against
			cumulative single-threaded LO time per pair. Scoring, sampling, and
			termination are shared and excluded; this is an accuracy--LO-cost frontier,
			not an end-to-end speed comparison. Beyond the smallest budget, Welsch-HQ retains the highest accuracy
			at matched LO time despite its higher per-call cost.}
		\label{fig:supp-budget-time}
	\end{figure}

	Local optimization is most valuable when the sample budget is small and
	the selected model is inaccurate. On easy ETH3D pairs, most selected models are
	already inside every tested basin, so the arms lie within 0.04 of each other, except TLS-LO$^+$.

	\paragraph{Timing protocol.}
We measure local-optimization time with a dedicated single-threaded harness
that replays the shared hypothesis sequence on one dense PhotoTourism scene
(40 pairs, about 4,000 correspondences). Each optimizer uses its deployed
settings and the shared 25-model-update cap. We divide accumulated LO time by
the number of calls, excluding workspace allocation and first-call
compilation. The \tlsirls{} and TLS-LZ refits are timed as separate columns in the
same run; each other optimizer is measured in its own run. These measurements
underlie the design table and accuracy-versus-LO-time comparison in the main
paper.

\small

\FloatBarrier
\end{document}